\documentclass{article}

\PassOptionsToPackage{numbers}{natbib}

\usepackage[final]{neurips_2024}

\usepackage{pifont}
\usepackage[utf8]{inputenc} 
\usepackage[T1]{fontenc}    
\usepackage{url}            
\usepackage{booktabs}       
\usepackage{amsfonts}       
\usepackage{nicefrac}       
\usepackage{microtype}      
\usepackage[table,usenames,dvipsnames]{xcolor}      
\usepackage{wrapfig}
\usepackage[colorlinks,linktoc=all]{hyperref}
\usepackage{tikz}
\usepackage{amsmath,amsfonts,mathtools}
\usepackage{graphicx}
\usepackage{arydshln}
\usepackage{amsmath}
\usepackage{algorithm}
\usepackage{algpseudocode}
\hypersetup{citecolor=Blue}
\hypersetup{linkcolor=Blue}
\hypersetup{urlcolor=Blue}

\usepackage{hyperref}
\usepackage{url}  
\usepackage{amsmath}
\usepackage{amssymb}
\usepackage{mathtools}
\usepackage{amsthm}
\usepackage{xspace}

\usepackage{amsmath,amsfonts,bm}

\def\eqref#1{equation~\ref{#1}}

\def\1{\bm{1}}

\DeclareMathAlphabet{\mathsfit}{\encodingdefault}{\sfdefault}{m}{sl}
\SetMathAlphabet{\mathsfit}{bold}{\encodingdefault}{\sfdefault}{bx}{n}

\def\gG{{\mathcal{G}}}

\usepackage[capitalize,noabbrev]{cleveref}

\theoremstyle{plain}

\theoremstyle{definition}

\theoremstyle{remark}

\usepackage[textsize=tiny]{todonotes}
\usepackage{multirow}
\usepackage{subcaption}

\usepackage{enumitem}

\usepackage{makecell}

\newcommand{\methodname}{{\textsc{RAGraph}}\xspace}

\newlength\myheight
\newlength\mydepth
\settototalheight\myheight{Xygp}
\title{
    \parbox{0.185\textwidth}{%
        \vspace{-0.63cm}
        \centering
        \includegraphics[height=3.6em]{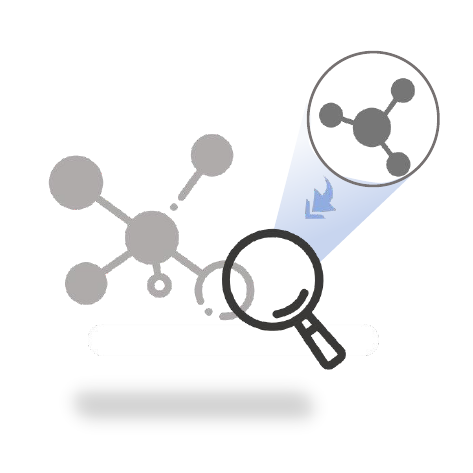}\\
        \vspace{-0.85cm}
    }%
    \textsc{RAGraph}: A General Retrieval-Augmented\\ \quad \quad  \quad Graph Learning Framework
}

\author{
  Xinke Jiang$^{\spadesuit *}$, Rihong Qiu$^{\spadesuit *}$, Yongxin Xu$^{\spadesuit}$\thanks{Indicates equal contribution.}~~, Wentao Zhang$^{\diamondsuit}$, Yichen Zhu$^{\diamondsuit}$,
  Ruizhe Zhang$^{\spadesuit}$  \\ \textbf{Yuchen Fang$^{\heartsuit}$, Xu Chu$^{\blacklozenge \bigstar}$, 
  Junfeng Zhao$^{\spadesuit \clubsuit}$}\thanks{Yasha Wang and Junfeng Zhao are corresponding authors.}~~, \textbf{Yasha Wang$^{\spadesuit \bigstar \dagger}$}  \\
   \includegraphics[height=0.83em]{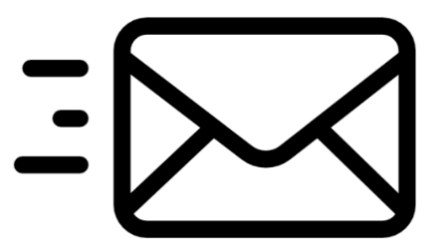} \texttt{\{xinkejiang, rihongqiu, xuyx, ruizhezhang\}@stu.pku.edu.cn} 
  \\
  \includegraphics[height=0.83em]{Graphs/email.png} \texttt{\{wentaozh2001, yichenzhu2014, fyclmiss\}@gmail.com} \\
  \includegraphics[height=0.83em]{Graphs/email.png} \texttt{\{chu\_xu, zhaojf, wangyasha\}@pku.edu.cn} \\
  \normalfont{$^\spadesuit$Key Laboratory of High Confidence Software Technologies (Peking University),} \\
  \normalfont{School of Computer Science, Peking University, China} \\
  \normalfont{$^\diamondsuit$No Affiliation, $^\heartsuit$University of Electronic Science and Technology of China} \\
  \normalfont{$^\blacklozenge$Center on Frontiers of Computing Studies, Peking University, Beijing, China} \\
  \normalfont{$^\clubsuit$Big Data Technology Research Center, Nanhu Laboratory, Jiaxing, China} \\
  \normalfont{$^\bigstar$Peking University Information Technology Institute, Tianjin Binhai, China} \\
}

\usepackage[subtle]{savetrees}

\begin{document}
\maketitle

\vspace{-1cm}
\begin{center}{
  \raisebox{-\mydepth}{\includegraphics[height=1.4\myheight]{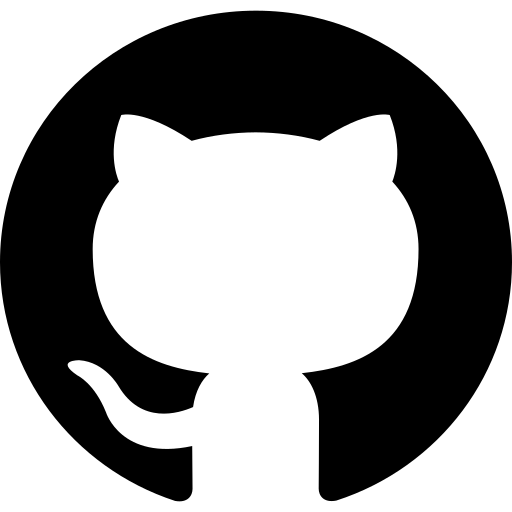}}
\textbf{\url{https://github.com/Artessay/RAGraph/}}
}
\end{center}

\begin{abstract}
Graph Neural Networks (GNNs) have become essential in interpreting relational data across various domains, yet, they often struggle to generalize to unseen graph data that differs markedly from training instances. In this paper, we introduce a novel framework called General \textbf{\underline{R}}etrieval-\textbf{\underline{A}}ugmented \textbf{\underline{Graph}} Learning (\textbf{\methodname}), which brings external graph data into the general graph foundation model to improve model generalization on unseen scenarios. On the top of our framework is a toy graph vector library that we established, which captures key attributes, such as features and task-specific label information. During inference, the \textbf{\methodname} adeptly retrieves similar toy graphs based on key similarities in downstream tasks, integrating the retrieved data to enrich the learning context via the message-passing prompting mechanism. Our extensive experimental evaluations demonstrate that \textbf{\methodname} significantly outperforms state-of-the-art graph learning methods in multiple tasks such as node classification, link prediction, and graph classification across both dynamic and static datasets. Furthermore, extensive testing confirms that \textbf{\methodname} consistently maintains high performance without the need for task-specific fine-tuning, highlighting its adaptability, robustness, and broad applicability.
\end{abstract}

\section{Introduction}
\textbf{Graph Neural Networks (GNNs)}~\cite{cai2018comp,GCN,GAT,liu2021pick,Zhang2024InfiniteHorizonGF} have recently burgeoned a surge of interest in both academic and industry communities due to their robust capability to model complex, real-world data in diverse domains, including societal~\cite{sn_IJCAL,aorec, qin2022next}, biochemical~\cite{duv2015conv,yang2023kerprint,xu2023seqcare}, and traffic-related~\cite{SPGCL, gao2023spatiotemporal, sttd, fang2023spatio, fang2024efficient} fields and etc~\cite{CAS2, CAS3,Luo2024TimeseriesSA,Duan2024CaTGNNEC,gao2024uncertainty,gao2023uncertainty}.
Utilizing a message-passing mechanism~\cite{GCN, hamilton2017inductive}, GNNs have transcended traditional node embedding approaches~\cite{grover2016node2vec, 2014deepwalk, tang2015line}, enabling the capture of intricate relationships within data through sophisticated architectures and advanced graph representation learning techniques~\cite{GCN, ResGCN, SPGCL, GPRGNN, GAT}.
However, 
the challenge of generalizing GNNs across different modalities, domains~\cite{liu2023flood,liu2022ud}, and tasks remains largely unexplored~\cite{li2024graphadapter, yang2024graphpro}. This is in stark contrast to the significant successes of large models such as GPTs~\cite{ChatGPT, OpenAI2023GPT4TR} in NLP and Sora~\cite{sora} in CV, presenting a crucial frontier for further research and realms for graph data generalizing. 

\begin{wrapfigure}{r}{10cm}
\vspace{-0.5cm}
\includegraphics[width=\linewidth]
{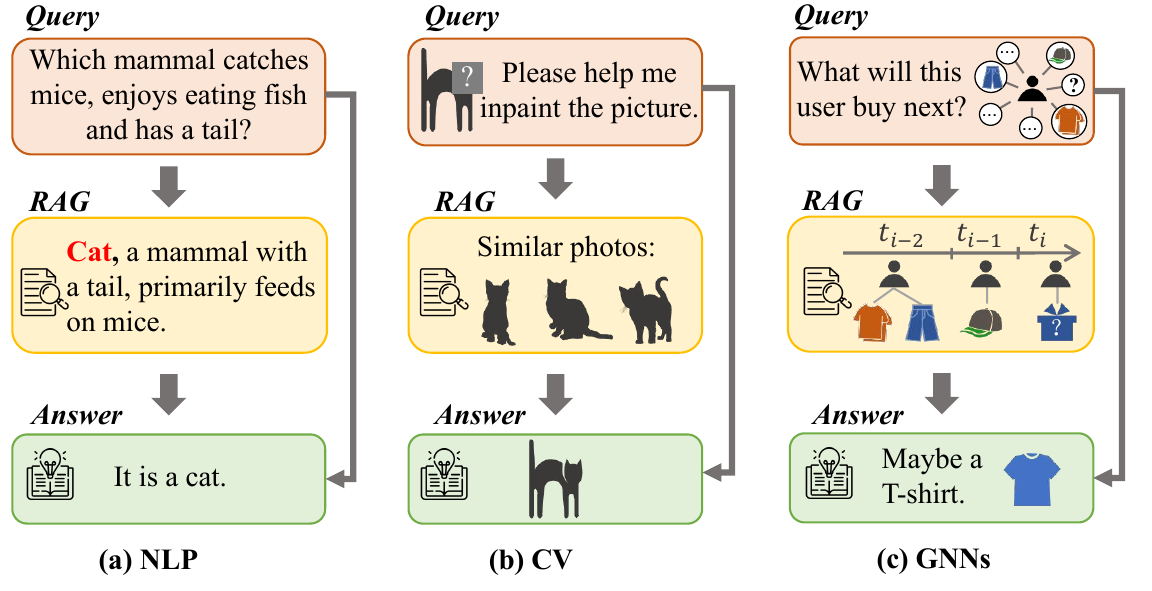}
\vspace{-0.45cm}
    \caption{\textbf{(a)} RAG in NLP utilizes retrieval to enhance model responses, based on a query to retrieve related features (\emph{e.g.,} \textit{a tail, primarily feeds on mice}) and answers (\emph{e.g.,} \textit{Cat}). \textbf{(b)} In CV, RAG employs similar photo retrieval to enhance model comprehension, assisting in downstream tasks such as inpainting or image question answering. \textbf{(c) }For GNNs, RAG could leverage retrieval of similar historical subgraphs or scenarios to aid in graph-based tasks (\emph{e.g.,} recommendations or fraud detection).}
    \vspace{-0.2cm}
    \label{fig:motiv_relu}
\end{wrapfigure}

In graph learning tasks, providing the necessary context is crucial for graph generalization~\cite{zhang2023graph,li2022outofdistribution,mishchenko2023prodigy,101}, \emph{i.e.,} retrieve similar shopping context as illustrated in Figure~\ref{fig:motiv_relu} (c). Therefore, our insight is to enhance the model's generalization ability and prediction accuracy by retrieving necessary contexts during graph learning through retrieval.
\textbf{Retrieval-Augmented Generation (RAG)} represents a prominent methodology, significantly augmenting language model functionalities through the integration of a dynamic retrieval mechanism during the generation process \citep{rag,parashar2024neglected} (\emph{e.g.}, a person asks what animal it is, and we use some visual~\cite{zhu2024retrievalaugmented} or text retrieval~\cite{asai2023selfrag} methods to retrieve more descriptive features or even the wanted category). 
RAG enriches not only accurate and reliable content but also reduces factual errors, addressing challenges such as incorrect answers, hallucinations, and limited interpretability in knowledge-intensive tasks \citep{izacard2022atlas, asai2023selfrag, asai2023retrieval}, obviating the need for updating model parameters and could be generalized even in unseen scenarios.

However, how to enable retrieval-augmented generation for graph learning, \emph{i.e.}, retrieving the user's historical purchasing behavior to enhance recommendation ability~\cite{he2020lightgcn, yang2024graphpro,huang2024recallaugmented} and identifying fraud crimes by searching for similar fraudulent relationship behaviors~\cite{misinfo, liu2021pick}, still remains unexplored and faces the following challenges \textit{\textbf{C1\& C2}}.

\noindent\textit{\textbf{C1.}} The first challenge is how to leverage the retrieved context \emph{i.e.,} features ($X$) and labels ($Y$) into the GNNs model under dynamic changing scenarios.
Previous studies, such as PRODIGY~\cite{mishchenko2023prodigy}, have adopted the concept of in-context learning (ICL) by constructing consistent and static task graphs for each specific task or dataset.
These task graphs determine labels through the calculation of similarities using hidden vectors, employing a few-shot learning approach. However, PRODIGY's reliance on a fixed set of examples as rules may not sufficiently address and generalize the variety of scenarios encountered in real-world settings, which is particularly problematic in dynamically changing environments, as the system focuses primarily on teaching the direct mapping paradigm from inputs to outputs ($X\to Y$), rather than truly integrate the input ($X$) and output ($Y$) data into the analysis. In contrast to RAG, PRODIGY struggles to incorporate external information ($X$ and $Y$) related to data nodes, which is crucial for enriching the learning process in graph-based systems.

\noindent\textit{\textbf{C2.}} Moreover, it is challenging to develop a tune-free prompt mechanism to support retrieved knowledge and be applicable to seamlessly switch unseen scenarios and multi-tasks. 
Numerous initiatives have been undertaken in the realm of graph pre-training~\cite{hu2019strategies,graphcl,hu2020gpt,qiu2020gcc,tent, gmeta, chen2019meta, sun2021one, zhang2020few}, however, the challenge persists in designing a plug-and-play RAG module that can seamlessly interface with already pre-trained models.
Insights derived from prior investigations into the graph prompt~\cite{chen2024prompt,gao2024protein,sun2023all,graphprompt,yang2024graphpro,fang2024universal,VNT}, the knowledge obtained by RAG can be facilitated and injected into prompt via a plug-and-play manner.

For endeavoring to address these two challenges previously mentioned, we put forward the General \textbf{\underline{R}}etrieval- \textbf{\underline{A}}ugmented \textbf{\underline{Graph}} Learning Framework (\textbf{\methodname}). Drawing inspiration from the success of RAG on LLMs~\cite{rag} and the ICL on GNNs~\cite{mishchenko2023prodigy} (we detail the difference between RAG and ICL in Appendix~\ref{sec: app diference}), we constructed a toy graphs vector library by chunking from resource graphs, where the library key stores key information, including environmental, historical, structural, and semantic details, while node features and label information (task-specific output vector) are stored as values. 
For downstream tasks, the key value of the query node would be leveraged to retrieve toy graphs by the key similarities, and the stored features (X) and labels (Y) would be aggregated structurally to provide essential knowledge to the query node, instead of the mapping paradigm, to address challenge \noindent\textit{\textbf{C1}}.
In prompt mechanism design, we start by transferring features and task-specific output from the toy graphs to their master nodes (the central node of the toy graph) via message-passing. Subsequently, features from the master nodes and the query node's neighbors are aggregated to the query node, along with the task-specific output from master nodes. This process could be parameter-free, indicating that our model can be applied across different tasks and datasets without the need to fine-tune for downstream tasks, effectively addressing challenge \noindent\textit{\textbf{C2}}.

In summary, our contributions are listed as follows:
\begin{itemize} [leftmargin=*,itemsep=0pt]

\item To the best of our knowledge, our proposed framework, \methodname, is the first to integrate RAG with pre-trained GNNs. By constructing a key-value vector library for toy graphs, \textit{\methodname} facilitates explicit plug-and-play access to pre-trained GNNs, achieving commendable performance even without fine-tuning, demonstrating its superiority on cross-task and cross-dataset capabilities.

\item Our \methodname employs a classic message-passing mechanism and introduces a well-designed prompt mechanism to integrate knowledge. This approach effectively incorporates the retrieved knowledge $X$ and $Y$ from toy graphs, into the pre-trained GNNs model, enhancing the accuracy and relevance of the model's outputs.

\item We have extensively tested \methodname on both static and dynamic graphs across multiple levels of graph tasks (node, edge, and graph). The results validate the effectiveness of our model, showing significant improvements over state-of-the-art baselines in both fine-tuned and tuning-free scenarios, particularly in cross-dataset validations.
\end{itemize}
\section{Related Work}

\subsection{Retrieval-Augmented Generation on Large Language Models}
RAG integrates an external knowledge retrieval component and through prompt engineering into pre-trained language models to enhance factual consistency, thus improving the reliability and interpretability of LLM responses~\cite{zhao2023retrieving, rag, lewis2021retrievalaugmented, gao2022precise, hykge, con, cok, xu2024parenting, zhang2024kapo}. Traditional RAG approaches utilize retriever models to source relevant documents from extensive knowledge corpora~\cite{xu2024retrieval, qu2021rocketqa, ma2023chainofskills, chen2024improving}, which are then processed further by reader models—primarily LLMs~\cite{kg_review, sarthi2024raptor}.
Furthermore, several studies focus on fine-tuning reader LLMs by applying prompt-tuning with retrieved knowledge or using RAG API calls~\cite{luo2023sail, izacard2022atlas, asai2023selfrag, yoran2023making, wang2024rear, zhang2023knowledgeable, lin2023radit}. While RAG has seen considerable success in the NLP field, it has also been applied to tasks involving joint visual and text retrieval~\cite{zhu2024retrievalaugmented, lin2024preflmr, lin2022retrieval, chen2022murag, yuan2023ramm,czt}, code retrieval~\cite{lu2022reacc, zhou2023docprompting}, audio retrieval~\cite{chan2023using,he2023alcap} and video retrieval~\cite{bogolin2022cross,wang2021t2vlad}.
Although there have been applications of RAG on structured data such as KG-RAG for knowledge graphs~\cite{jin2023heterformer, hykge, kgrag_arxiv,tan2024musegraph,NEURIPS2023_2ac879d1,huang2022fewshot}, these primarily leverage the text information of knowledge graph nodes to enhance language or graph models. In contrast, there are no significant studies utilizing RAG on structured graphs without text information to enhance pre-trained GNNs.
Our work aims to extend this successful approach similarly to graph data, to enhance the capabilities of pre-trained GNNs, and can be adapted to various tasks and across different graphs without additional fine-tuning by integrating a plug-and-play RAG module.

\subsection{Graph Prompt Learning}
Inspired by the application of pre-training models~\cite{ChatGPT,OpenAI2023GPT4TR} and prompt learning~\cite{wei2023chainofthought,zhou2023docprompting,Jiang2024TCRAGTuringCompleteRC} in NLP, recently, learning on the graph has been divided into pre-training models on large-scale graph data~\cite{hu2019strategies,graphcl,hu2020gpt,qiu2020gcc,mishchenko2023prodigy,zhao2024all,xia2024opengraph,gppt,yu2024few,yu2024dygprompt,yu2024non,yu2023generalized,
yu2023multigprompt}, with or without labels, followed by fine-tuning model parameters via prompts for diverse downstream tasks~\cite{graphprompt,yang2024graphpro,mishchenko2023prodigy,zhu2023sglpt,gppt,VNT}. 
The adoption of prompting mechanisms in graph learning represents a promising avenue to overcome the constraints of traditional graph representation methods, striking a balance between flexibility and expressiveness~\cite{sun2023graph}. For instance, VNT~\cite{VNT} utilizes virtual nodes as prompts to refine the application of pre-trained graph models. GraphPrompt~\cite{graphprompt} introduces a task-specific readout mechanism to tailor models for various tasks, while GraphPro~\cite{yang2024graphpro} implements spatial- and temporal-based gating mechanisms suited for dynamic recommendation systems.
Furthermore, PRODIGY~\cite{mishchenko2023prodigy} constructs task graphs (prompts) and data graphs to enhance the model's ICL capabilities. 
Leveraging the successes in graph prompt learning, we aim to inject retrieved knowledge via prompt into pre-trained GNNs to support downstream tasks.

\section{Preliminaries}
In \methodname, we focus on RAG on multi-level graph tasks.
For consistency, we define the graphs as dynamic graphs, considering static graphs as the special cases within this framework. 
The subsequent definition provides a detailed description of toy graphs, including the definitions of keys and values utilized in \methodname. Additionally, inspired by GraphPrompt~\cite{graphprompt}, we have unified node-level, edge-level, and graph-level tasks into a cohesive framework, and employ query graphs to tackle downstream tasks with precision.

\paragraph{\textit{Definition 1. (Dynamic Graph)}}{
Let $\mathcal{G}=\{G_t\}_{t=1}^T$ denote a dynamic graph comprising a sequence of graph snapshots, each represented as a static graph $G_t= (V_t, E_t, X_t, A_t, Y_t)$. $\mathcal{V} = \bigcup_{t=1}^T V_t = \{v_1, ..., v_n\}$ defines the combined set of nodes across all snapshots and $\mathcal{E} = \bigcup_{t=1}^T E_t \subseteq \mathcal{V} \times \mathcal{V}$ is the edge set,
where $V_t$ and $E_t$ represent the nodes and edges of the $t$-th snapshot, respectively.
Feature matrix $X_t = \{x_v \mid v \in \mathcal{V}\} \in \mathbb{R}^{n \times d}$ contains the feature vectors for the nodes in the $t$-th snapshot, where $d$ is the feature dimension. $A_t$ denotes the edge weight matrix at time $t$, 
where edge weight $A_{t}[i,j] \in (0,1]$ if $v_i,v_j \in V_t$ and $(v_i,v_j) \in E_t$, and 0 otherwise. Furthermore, $Y_t$ represents the task-related labels associated with nodes, edges, or the graph at time $t$. Note that a graph is static if $T=1$ and for consistency in terminology, we unify static graphs as a particular instance of dynamic graphs.

\paragraph{\textit{Definition 2. (Toy Graph Vector Base)}}{Let \(\mathcal{G^R} = \{G_t^\mathcal{R}\}_{t=1}^T\) denote a dynamic resource graph. We chunk $\mathcal{G^R}$ into snapshots and take each node in $\mathcal{G^R}$ as the master node $v_m$ of the corresponding toy graph, and then store $v_m$ with its neighbors within $k$ hops as subgraphs. Data augmentation techniques~\cite{SPGCL,zhou2023data} such as node dropout, edge dropout, and random noise addition are employed on subgraphs to enhance the robustness and variability when generating each toy graph $G^\mathcal{T}$ (c.f. Section~\ref{sec:Toy Graphs Augmentation Strategy} for details). 
Each toy graph \(G^\mathcal{T} \subseteq \mathcal{G^R}\) is associated with a specific timestamp \(\tau\) and master node $v_m \in \mathcal{V}$, with each toy graph's scale being considerably smaller in scale compared to their corresponding $\mathcal{G^R}$. 
\textbf{\ding{182}} Toy graphs can be retrieved using \textbf{keys} that include the timestamp \(\tau\), the hidden embedding of the master node $h^{\tau}_m \in \mathbb{R}^{f_1}$ (\emph{e.g.,} embedded by pre-trained GNNs in \methodname), the environmental key (\emph{e.g.,} the neighbors set $\mathcal{N}(v_m^{\tau}) = \{v_i^{\tau} \mid A_{\tau}[m,i] > 0, v_i^{\tau} \in G^\mathcal{T} \}$) and the structure-based position-aware code $s^{\tau}_m$ ({cf.} Appendix~\ref{sec:key construction} for details). \textbf{\ding{183}} By retrieving based on key similarity ({c.f.} Section~\ref{sec:Toy Graphs Retrieval Process} for details), we can obtain the required \textbf{values} of $G^\mathcal{T}$, \textit{i.e.} task-specific output vector
$\{o_i^\tau \in \mathbb{R}^{f_2} | v_i \in G^{\mathcal{T}}\}$ and hidden embeddings $\{h_i^\tau \in \mathbb{R}^{f_1} | v_i \in G^{\mathcal{T}}\}$ of the master node and its neighbors, where $f_1$ and $f_2$ represent the dimensions. Finally, we denote the \textbf{key-value} vector base for the toy graph as $\mathcal{G^T}$.

\paragraph{\textit{Definition 3. (A Unified Graph Task Definition)}} Given a dynamic graph $\mathcal{G}$, it can be divided into training and testing subsets, \emph{i.e.} $\mathcal{G} = \mathcal{G}_{\rm{train}} \cup \mathcal{G}_{\rm{test}}$ based on either snapshot or node set partitioning. The label $y_i$ of a node $v_i$, edge $(v_i,v_j)$ or subgraph $G_i$ can be observed only if they belong to $\mathcal{G}_{\rm{train}}$. The objective of label prediction is to predict test labels $Y_{\rm{test}} \in \mathcal{G}_{\rm{test}}$. Following GraphPrompt~\cite{graphprompt}, we unify the three types of graph learning tasks (node-level, edge-level, and graph-level) into a single framework via similarity comparison $\texttt{sim}(\cdot,\cdot)$ of the task-specific output vector (abbreviated as $O$, where each entry is $o$) with the ground-truth (\textit{i.e.,} the one-hot vector or the prototype embedding under few-shot setting). It's noted that $o$ can be either low-dimensional (with the dimension equal to the number of predicted classes) under normal settings~\cite{GCN,Zhang2024InfiniteHorizonGF}, or high-dimensional under few-shot settings~\cite{graphprompt} or in link prediction tasks~\cite{yang2024graphpro,he2020lightgcn}.
In our experiment, \ding{182} for node-level and graph-level tasks, the downstream tasks are given in few-shot settings following ~\cite{graphprompt}: For node~/~graph classification on a node~/~graph set, let $\mathcal{C}$ be the set of classes with $y_i\in \mathcal{C}$ denoting the class label of node~/~graph. For each node~/~graph class, the class prototypical output vector is calculated by the mean value of the $\kappa$-shot set $\mathcal{D}$: ${\tilde{o}}_c = \frac{1}{\kappa}\sum_{(i,y_i)\in \mathcal{D}, y_i=c} {o}_{i}$. 
The class $y_i$ of the node or graph is determined by calculating similarity with the class prototype as:
\(
y_i = \arg\max_{c \in \mathcal{C}} \texttt{sim}({o}_{i}, {\tilde{o}}_c)
\).
\ding{183} For edge-level tasks, to predict a link between nodes $v_i$ and $v_q$, if $\exists v_j, (v_i,v_j)\in \mathcal{E}_{\text{train}}\in \mathcal{G}_{\text{train}}$ and $\texttt{sim}(o_i,o_q) \geq \texttt{sim}(o_i,o_j) + \epsilon$, we regard $(v_i,v_q)$ as linked. 
Following PRODIGY~\cite{mishchenko2023prodigy} and GraphPrompt~\cite{graphprompt}, we also apply a query graph \(G^\mathcal{Q}\) that includes the center node and its neighbors within \(k\) hops. 
Specifically, for graph-level task, we apply a full-link virtual node as the center node inside the query graph \(G^\mathcal{Q}\). 
\section{\methodname Framework}
In this section, we introduce \methodname, a general and novel retrieval-augmented graph learning framework that can operate on arbitrary graphs {with or without additional fine-tuning}, as illustrated in Figure~\ref{fig:main_figure}. Initially, in Section~\ref{sec:Toy Graphs Embedding Pipeline}, we elucidate the methodology for constructing the Resource Toy Graphs. Subsequently, in Section~\ref{sec:Toy Graphs Retrieval Process} we detail the Toy Graphs Retrieval Process. Finally, the Training and Inference processes are elaborated in Section~\ref{sec:Training and Inference}, which utilize retrieved toy graphs from two propagation views—intra and inter-propagation—and handle two types of information: hidden embeddings and task-specific output vectors in two techniques (noisy trainable approach or parameter-free approach). 
The main notations of \methodname are summarized in Table \ref{tab:symbols}, Appendix~\ref{sec:app notation}. 
For enhanced clarity, the Toy Graph Construction is outlined in Algorithm~\ref{alg.toygraph} (cf. Appendix~\ref{app:sec alg}) and the Training and Inference with Toy Graphs Retrieval are detailed in Algorithm~\ref{alg.model} (cf. Appendix~\ref{app:sec alg}). Moreover, in Appendix~\ref{sec: app effectiveness of rag}, we theoretically prove the effectiveness of applying RAG on GNNs from the perspective of mutual information gain.

\begin{figure}[tbh!]
    \centering
    \includegraphics[width = 1.0\linewidth]{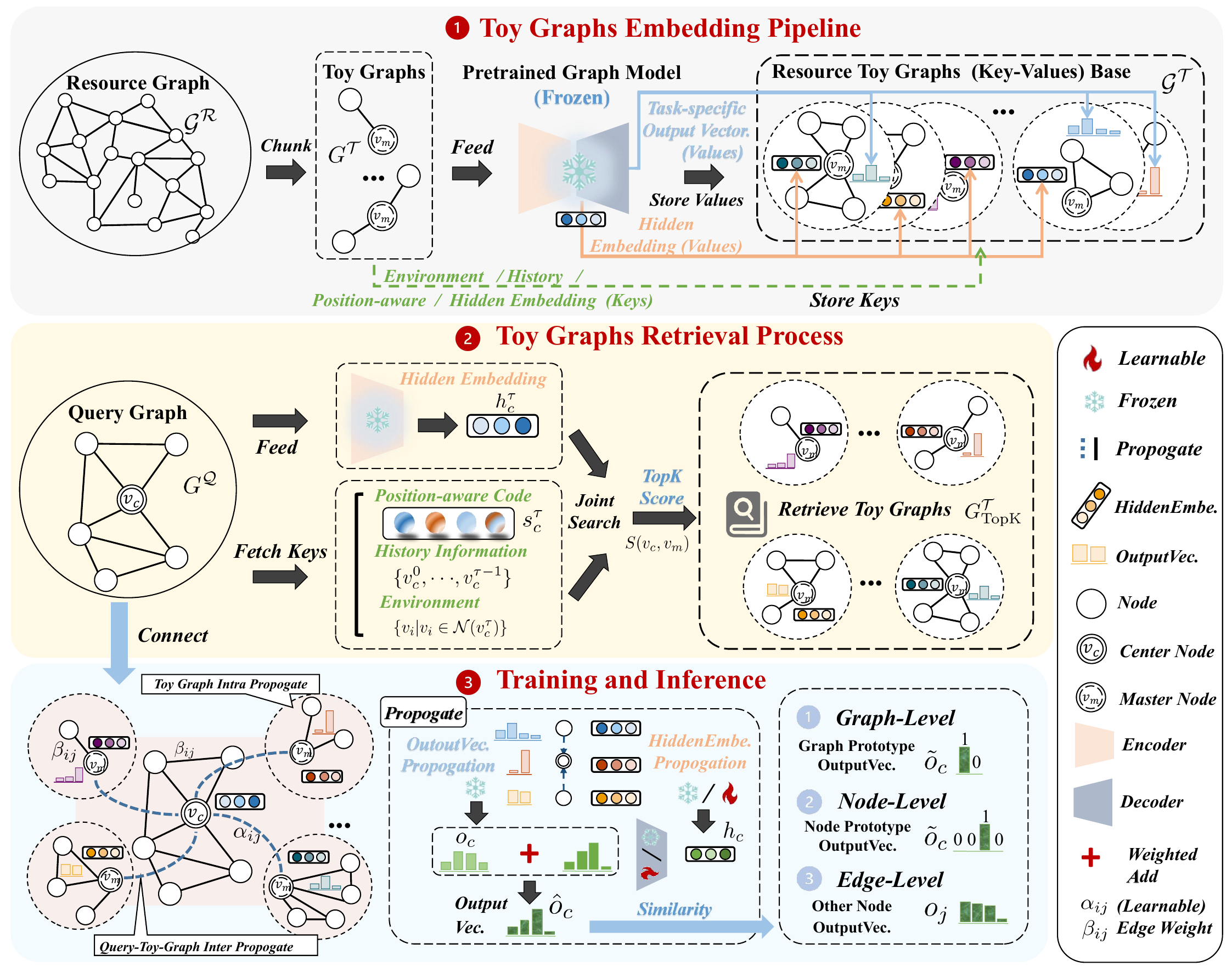}
    \caption{The overall framework of \methodname.
    \ding{182} Given resource graph $\gG^{\mathcal{R}}$, 
    we chunk it and augment toy graphs $\{G^{\mathcal{T}}\}$, and feed them into pre-trained GNNs to generate hidden embeddings via the encoder and task-specific output vectors via decoder, which are stored as values. Keys such as environment, history, position-aware, and hidden embeddings are stored to form the key-value database of toy graphs $\mathcal{G}^{\mathcal{T}}$. 
     \ding{183} For a given query graph $G^{\mathcal{Q}}$, the keys are fetched to retrieve the $topK$ toy graphs $G_{\text{topK}}^\mathcal{T}$ from the database.
     \ding{184} Leveraging $G_{\text{topK}}^\mathcal{T}$, intra- and inter-propagation are performed to propagate hidden embeddings and task-specific output vectors to pass retrieved knowledge to center node $v_c$. Through a weighted fusion, the aggregated output is used to perform graph-, node- and edge-level tasks. 
    }
    \label{fig:main_figure}
\end{figure}

\subsection{Toy Graphs Embedding Pipeline}
\label{sec:Toy Graphs Embedding Pipeline}
In graph-based learning, nodes with higher connectivity—typically with higher degrees—often hold more significance, meaning their information is more extensively learned during graph-pre-training processes. Conversely, less important nodes—those in the long tail—often have their features overlooked. 
This issue is particularly pronounced in LLMs performing RAG, where the predominance of common knowledge overshadows the long-tail knowledge that RAG is meant to leverage. To tackle this, we construct toy graphs using an inverse importance sampling strategy, thereby countering this bias by preferentially sampling and augmenting toy graphs that accentuate the long-tail knowledge. 

\paragraph{\textbf{Inverse Importance Sampling Strategy.}} To achieve this, we calculate each node's importance $I(v)$ for node $v \in G^\mathcal{R}_\tau$ by combining PageRank $\textsc{PR}(v)$ and Degree Centrality $\textsc{DC}(v)$ using the formula \( I(v) = \alpha \text{PR}(v) + (1 - \alpha) \text{DC}(v) \), where \( \alpha \in (0,1) \) is the balance weight. We reverse the node importance with \( I'(v) = \frac{1}{I(v) + \epsilon}, \epsilon \to 0 \), normalize it to obtain node $v_i$ 's sampling probabilities \( p_i = \frac{I'(v_i)}{\sum_{j=1}^n I'(v_j)} \), and perform weighted sampling function \textsc{WeightedSampling}($G_\tau^{\mathcal{R}}$, $p_i$) to prioritize nodes with higher sampling probability (lower importance) according to $p_i$. 
When sampling, for each master node $v_m$, we generate its $k$-hop neighbors, termed an ego net \(G^e_{\tau}(v_m)\).
Given the constrained size of the resource graph, we adopt data augmentation techniques commonly used in contrastive learning~\cite{SPGCL,simgcl,graphcl} to enhance the representativeness and diversity of the resultant toy graphs.

\paragraph{\textbf{Toy Graphs Augmentation Strategy.}}
\label{sec:Toy Graphs Augmentation Strategy}
For augmentation, we first calculate the average reversed importance \(\bar{I'}(G^e_\tau(v_m))\) of the nodes within an ego graph as \( \bar{I'}(G^e_\tau(v_m)) = \frac{1}{|G^e_\tau(v_m)|} \sum_{v \in G^e_\tau(v_m)} I'(v) \), which then determines the number of augmentations \( n_{\text{aug}}(G^e_\tau(v_m)) = \lfloor K \cdot \bar{I'}(G^e_\tau(v_m)) \rfloor \), where \( K \) is a scaling constant that adjusts the intensity of the augmentation.
For node $v_i,v_j \in G^e_\tau(v_m)$, the augmentation techniques \textsc{DataAugmentation}($G^e_\tau(v_m)$, $n_{\text{aug}}$) employed include:
\begin{itemize}[leftmargin=*,itemsep=0pt]
    \item \textbf{\ding{182} Node Dropout:} \(v_i \in G^e_\tau(v_m)\) has a probability of being dropped: \( {p}(v_i \text{ being dropped}) = 1 - p_i \).
    \item \textbf{\ding{183} Addition of Gaussian Noise:} we add gaussian noise to node features as augmentation \(X'(v_i) = X(v_i) + \mathcal{N}(0, \sigma^2)\).
    \item \textbf{\ding{184} Node Interpolation:} a new node feature \(X'(v_{new})\) is created by linearly combining the features of two existing nodes \(v_i\) and \(v_j\), calculated as \(X'(v_{new}) = \lambda X(v_i) + (1 - \lambda) X(v_j), v_i, v_j \in G^{\mathcal{T}}\). And the edge weight between the new node \(v_{new}\) and node \(v_i\) is updated to $\lambda A[i,j]$ and node $v_j$ is $(1 - \lambda)A[i,j]$ accordingly~\cite{Xu2024ProtoMixAH}.
    \item \textbf{\ding{185} Edge Rewriting:} we alter connections based on the average of the involved nodes' sampling probabilities, expressed as \( p((v_i, v_j) \text{ being rewired}) = \frac{p_i + p_j}{2} \).
\end{itemize}

\paragraph{\textbf{Key-Value Pairs Construction}.} After completing the sampling and augmentation procedures, the generated toy graphs are transformed into key-value pairs for storage~\cite{xu2023vecocare}. Specifically, we collect each master node's $v_m$ historical information (timestamps $\tau$), environmental information (neighbors $\mathcal{N}(v_m^\tau)$), structural encodings $s_m^\tau$ (as described in the Appendix~\ref{sec:key construction}), and the hidden embeddings $h_m^\tau$ (obtained by processing the toy graph through the frozen pre-trained GNNs) and store them as keys at the master node $v_m$ of the toy graph. Additionally, we store task-specific output vectors $\{o_i^\tau | v_i \in G^{\mathcal{T}}\}$ and hidden embeddings $\{h_i^\tau | v_i \in G^{\mathcal{T}}\}$ as values at each node of the toy graph. 
For storage of these key-value pairs, we utilize the FAISS vector library~\cite{douze2024faiss} to facilitate accelerated retrieval and storage.

\subsection{Toy Graphs Retrieval Process}
\label{sec:Toy Graphs Retrieval Process}
After constructing the key-value toy graphs vector database, we proceed with the retrieval process for sub-tasks according to the four sub-similarities between the key values of the master node \(v_m\) in the toy graph and the center node \(v_c\) in the query graph, as detailed in Appendix~\ref{sec:app sim func}. The final similarity score is a weighted combination of these factors, and the \(topK\) toy graphs are selected as the retrieval results:
\begin{equation}
\label{eq:weighted_similarity}
S(v_c, v_m) = \mathbf{w} \times [S_{\text{time}}(v_c, v_m), S_{\text{structure}}(v_c, v_m), S_{\text{environment}}(v_c, v_m), S_{\text{semantic}}(v_c, v_m)]^\text{T},
\end{equation}
where $\mathbf{w}=[w_1, w_2, w_3, w_4]$ are the hyper-parameterized weights attributed to the time, structure, environment, and semantic similarities, respectively.
Using this composite similarity, we rank and retrieve the \(topK\) toy graphs:
\begin{equation}
G^\mathcal{T}_{\text{TopK}} = \text{TopK}_{G^{\mathcal{T}} \in {\mathcal{G^T}}} S(v_c, v_m),
\end{equation}
where \(G^\mathcal{T}_{\text{TopK}}\) represents the subset of toy graphs that best match the query based on the combined criteria.
This process ensures that we retrieve the most relevant toy graphs based on a comprehensive similarity measure, incorporating historical, structural, and environmental information. 

\subsection{Training and Inference}
\label{sec:Training and Inference}
In Section~\ref{sec:Knowledge Injection Propagation}, we detail the Knowledge Injection Propagation process, which includes two distinct propagation manners. Next, in Section~\ref{sec:Knowledge Fusion Layer}, we present our approach for combining the retrieved hidden embeddings with the task-specific output vectors.
Additionally, to enhance the robustness of \methodname, a noise-based prompt tuning strategy is introduced in Section~\ref{sec:ngpt}.

\subsubsection{Knowledge Injection Propagation}
\label{sec:Knowledge Injection Propagation}
After retrieving the \( topK \) toy graphs \( G^{\mathcal{T}}_{\text{TopK}} \), knowledge, specifically the task-specific output vectors \( O \) and hidden embeddings \( H \), is propagated from these toy graphs to the master nodes ({Toy Graph Intra Propagation}) and then to the center node \( v_c \) ({Query-Toy Graph Inter Propagation}). This propagation utilizes message-passing mechanisms via GNNs (cf. Appendix~\ref{sec: app revisiting GNN}). Each master node \( v_m \) in the toy graphs is connected to the center node \( v_c \) of the query graph based on the similarity scores \( S(v_c, v_m) \) computed in Eq.(\ref{eq:weighted_similarity}) and the connection weights dictate the influence of each toy graph, ensuring that graphs with higher similarity have a more substantial impact. This process can be implemented using either a parameter-free or a learnable approach. Moreover, it is worth noting that for learnable methods, the parameters of GNN are different.

\paragraph{\textbf{\ding{182} Toy Graph Intra Propagation}}{
Within each toy graph, information \( \mathbf{z} \) is propagated from neighbors to the master node using pre-trained GNNs. The task-specific output vectors \( o \) and hidden embeddings \( h \) from the neighbors are aggregated and transmitted to the master node. For each node \( v_i \) in a toy graph \( G^{\mathcal{T}} \), the GNN aggregates information from its neighbors \( \mathcal{N}(v_i) \) to update the master node \( v_m \):
\begin{equation}
\mathbf{z}_m = \textsc{Gnn}\left( \{\mathbf{z}_i \mid v_i \in \mathcal{N}(v_m) \} \right),
\end{equation}
where \( \mathbf{z}_i \) and \( \mathbf{z}_m \) represent the hidden embeddings \( h_i, h_m \) or task-specific output vectors \( o_i, o_m \) of the neighbor nodes and master node, respectively. For parameter-free situations, we can prepare $\mathbf{z}_m$ in advance when constructing the toy graph to improve inference efficiency.

\paragraph{\textbf{\ding{183} Query-Toy Graph Inter Propagation}}
Next, information from the toy graphs is aggregated to the query graph. Specifically, during propagation, information \( \mathbf{z} \) from the neighbors and master node of the toy graph is propagated to the center node using the same pre-trained GNNs. For a center node \( v_c \) in the query graph \( G^{\mathcal{Q}} \), the GNN aggregates hidden embeddings \( H \) from its neighbors \( \mathcal{N}(v_c) \) and the master node \( v_m \) from the toy graph:
\begin{equation}
h_c = \textsc{Gnn}\left( \{h_i \mid v_i \in \mathcal{N}(v_c) \cup \{v_m\} \} \right).
\end{equation}
When propagating the task-specific output vector \( O \), only the master node's information is passed to the center node:
\begin{equation}
o_c = \textsc{Gnn}\left( \{o_i \mid v_i \in \{v_m\} \} \right).
\end{equation}
For scenarios where the propagation mechanism is learnable, attention mechanisms can be adapted on the edges. In parameter-free scenarios—where there are no learnable weights—the attention on the edges is determined based on the edge weights from the previous resource graph.

\subsubsection{Knowledge Fusion Layer}
\label{sec:Knowledge Fusion Layer}
Finally, at the data fusion layer, the aggregated hidden embeddings \( H \) of the center node $v_c$ are processed through the pre-trained GNN's decoder \(\textsc{Decoder}(\cdot)\) to obtain an output vector \( O \). This output vector is then combined with the aggregated task-specific output vector in a weighted manner to produce the final output for downstream tasks as illustrated in Definition 3. The combined output is formulated as follows:
\begin{equation}
\hat{o}_{c} = \gamma o_c + (1-\gamma) \textsc{Decoder}(h_c),
\end{equation}
where \(\gamma\) is a reweighting hyper-parameter. The resulting vector \(\hat{o}_{c}\) is then utilized to perform node-, graph-, or edge-level tasks via a similarity function.

For the same task, the decoder can be directly used to generate outputs. For different tasks, the decoder can be masked, allowing the model to utilize pre-computed embeddings without additional training. Furthermore, the decoder can be fine-tuned to better meet the specific requirements of each task, providing both flexibility and optimized performance. This approach ensures that the model effectively integrates and leverages information from both the toy graphs and the query graph, enhancing its effectiveness in various downstream tasks through the use of the aggregated task-specific output vector.

\subsubsection{Noise-based Graph Prompting Tuning}
\label{sec:ngpt}
When prompt tuning, \methodname employs the same prompt loss function $\mathcal{L}_{\text{prompt}}$ as the backbone model (e.g., GraphPro, GraphPrompt). However, to mitigate the challenge of noise retrieval—a common issue in traditional RAG where highly related but irrelevant data is often retrieved—we enhance the training process by incorporating noise data to bolster model robustness, motivated by~\cite{CAS2}. Specifically, we implement two types of noise integration strategies:
\begin{itemize}[leftmargin=*,itemsep=0pt]
    \item \textbf{\ding{182} Inner-Toy-Graph Noise:} This strategy involves artificially introducing irrelevant nodes ($v_j \notin G^e_\tau(v_m)$) into the toy graph during its construction, complementing other augmentation techniques.
    \item \textbf{\ding{183} Toy-Graph Noise:} Throughout the training phase, we not only retrieve the \(topK\) toy graphs that are most relevant but also deliberately include the \(bottomK\) toy graphs to incorporate noise knowledge.
\end{itemize}
The integration of these noise elements is intended to enhance the model's ability to distinguish relevant information from irrelevant information, significantly improving its robustness and overall performance in downstream tasks by noise training. However, during the inference stage, we do not incorporate the noise.

\section{Experiments}
\label{sec:experiments}
In this section, we conduct a series of experiments to evaluate the performance of \methodname against state-of-the-art baselines on three dynamic and five static datasets on three-level graph tasks. Further details and experiment results are provided in Appendix~\ref{sec: Further Experiment Details}.

\subsection{Experimental Setup}
\label{sec:exp setup}

\paragraph{Datasets.} We use four static datasets \textit{PROTEINS, COX2, ENZYMES} and \textit{BZR} for graph classification and node classification, as well as three dynamic datasets \textit{TAOBAO, KOUBEI} and \textit{AMAZON} for link prediction. More details about these datasets can be found in Table~\ref{table:dataset statistics} in Appendix~\ref{appendix: datasets}. 

\paragraph{Methods and Baselines.}
We consider three versions of our proposed framework \methodname: 1) \methodname/NF, which indicates we utilize the plug-and-plag \methodname without fine-tuning on the train set; 2) \methodname/FT, which employs prompt tuning on the train set with RAG; and 3) \methodname/NFT, which applies
noise prompt tuning on the train set with RAG. 
For the baseline of the dynamic graph, we choose LightGCN~\cite{he2020lightgcn}, SGL~\cite{sgl}, MixGCF~\cite{huang2021mixgcf}, SimGCL~\cite{simgcl}, GraphPro~\cite{yang2024graphpro} and GraphPro+PRODIGY~\cite{mishchenko2023prodigy}. For the static graph, we choose GCN~\cite{GCN}, GraphSAGE~\cite{hamilton2017inductive},GAT~\cite{GAT}, GIN~\cite{gin}, GraphPrompt~\cite{graphprompt}, GraphPrompt+PRODIGY~\cite{mishchenko2023prodigy} as baselines. 
In addition, we denote '/NF' and '/FT' respectively to represent without fine-tuning and fine-tuning.
A detailed description of baselines can be referred to in Appendix~\ref{sec:appendix baselines}. 

\paragraph{Settings and Evaluation.}
We establish a training-resource split with the remainder of the data reserved as unseen during fine-tuning. For static graphs, the split is based on node partitioning with the ratio of 50\%:30\%~\cite{graphprompt}, while for dynamic graphs, it is based on partitioning snapshots with the history snapshots as resource graph~\cite{mishchenko2023prodigy}.
For fair comparisons, for methods employing PRODIGY and \methodname, \ding{182} we fine-tune models using the training set while retrieving the resource graph to prevent information leakage and over-fitting; \ding{183} when testing, we retrieve the combined training and resource graphs. For other methods, fine-tuning was directly performed on the combined train and resource set for fairness.
For the evaluation of static graphs, we refer GraphPrompt, utilizing pre-trained GNNs for both node- and graph-level tasks within a \(k\)-shot classification framework. For dynamic graphs, we follow GraphPro to employ pre-trained GNNs on a substantial dataset fraction, with fine-tuning and testing conducted on later snapshots.
Moreover, we pre-train GraphPro and GraphPrompt unsupervised on other datasets within the similar domain following~\cite{graphprompt,mishchenko2023prodigy} to avoid information leakage.
For classification tasks, we utilize the accuracy as evaluation matric; For link prediction tasks, we use standard metrics Recall@k and nDCG@k at $k=20$, in line with existing methodologies \cite{he2020lightgcn, sgl, simgcl}. The metrics used in the experiment are detailed in Appendix~\ref{appendix:evaluation matrics} and the implementation details of \methodname and baselines are in Appendix~\ref{app.setting-parameter}.

\subsection{Retrieval-Augmented Graph Results}
\label{sec:expt:perf}
As discussed, we conduct experiments and report the results of the three graph tasks for static graph and dynamic graph, as illustrated in Table~\ref{table.combined-classification} and Table~\ref{tab:overall}. From the reported accuracy, we can find the following observations:

\paragraph{Outperforming SOTA Methods.}
First, our proposed \methodname outperforms almost all the baselines across the three graph tasks, demonstrating the effectiveness of \methodname in transferring knowledge from the pre-training to downstream tasks compared to traditional GNNs \emph{i.e.,} GCN and GraphSAGE. It achieves the highest average accuracy across almost all tasks on ENZYMES, with an improvement of at least 5.19\% in the static graph, and up to 11.78‰ on the dynamic graph over the best baseline PRODIGY/FT. We argue that by virtue of the integration of hidden embedding and task-specific output vector, \methodname is able to comprehend more knowledge than simply learns the paradigm from $X\to Y$. Second, compared with the models of PRODIGY/NF and \methodname/NF, the introduction of noise training in noise prompt tuning also improves the robustness of the model, avoiding the influence of a large amount of noise on the information aggregation inside the query graph.

\paragraph{Strong Retrieval-Augmented Performance on Unseen Datasets.} 
\begin{wrapfigure}{r}{8cm}
\vspace{-0.5cm}
\includegraphics[width=0.47\linewidth]{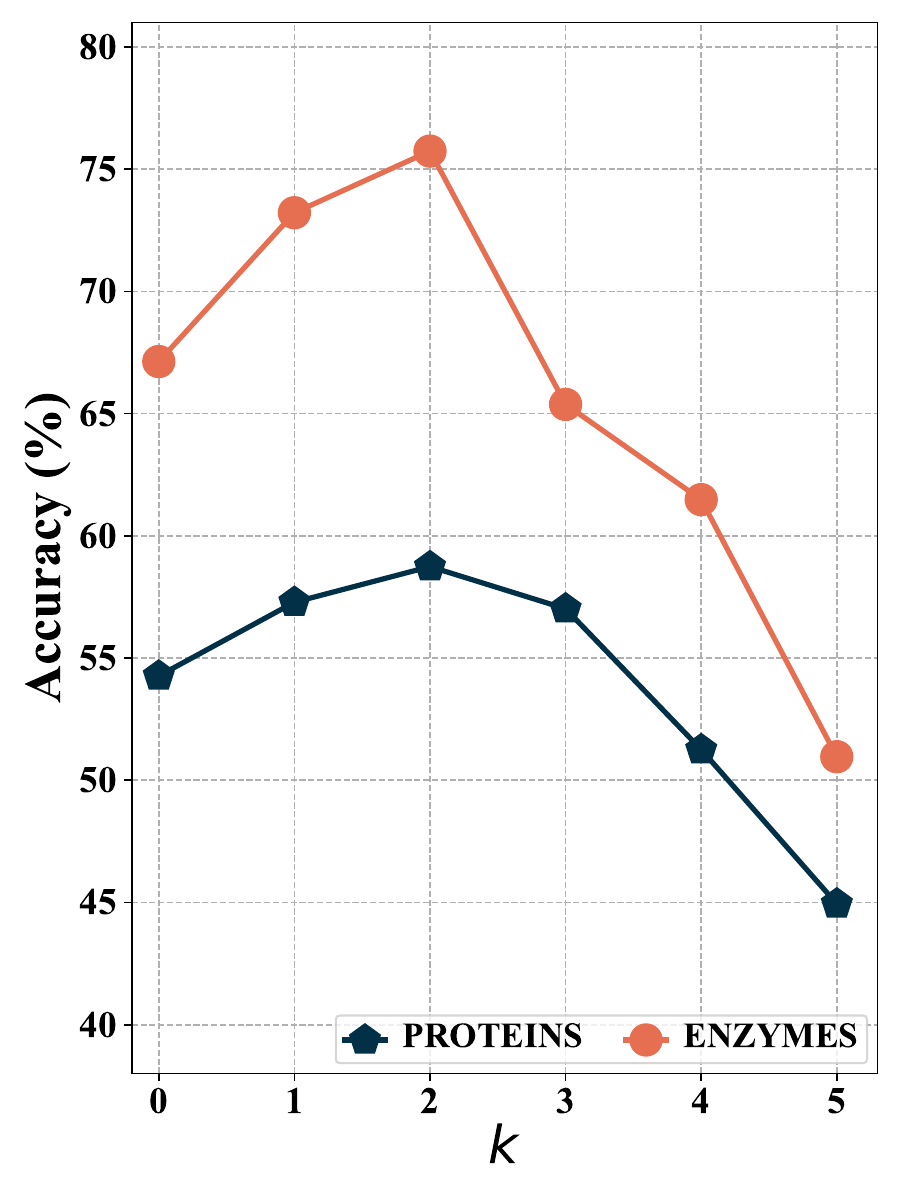}
\includegraphics[width=0.47\linewidth]{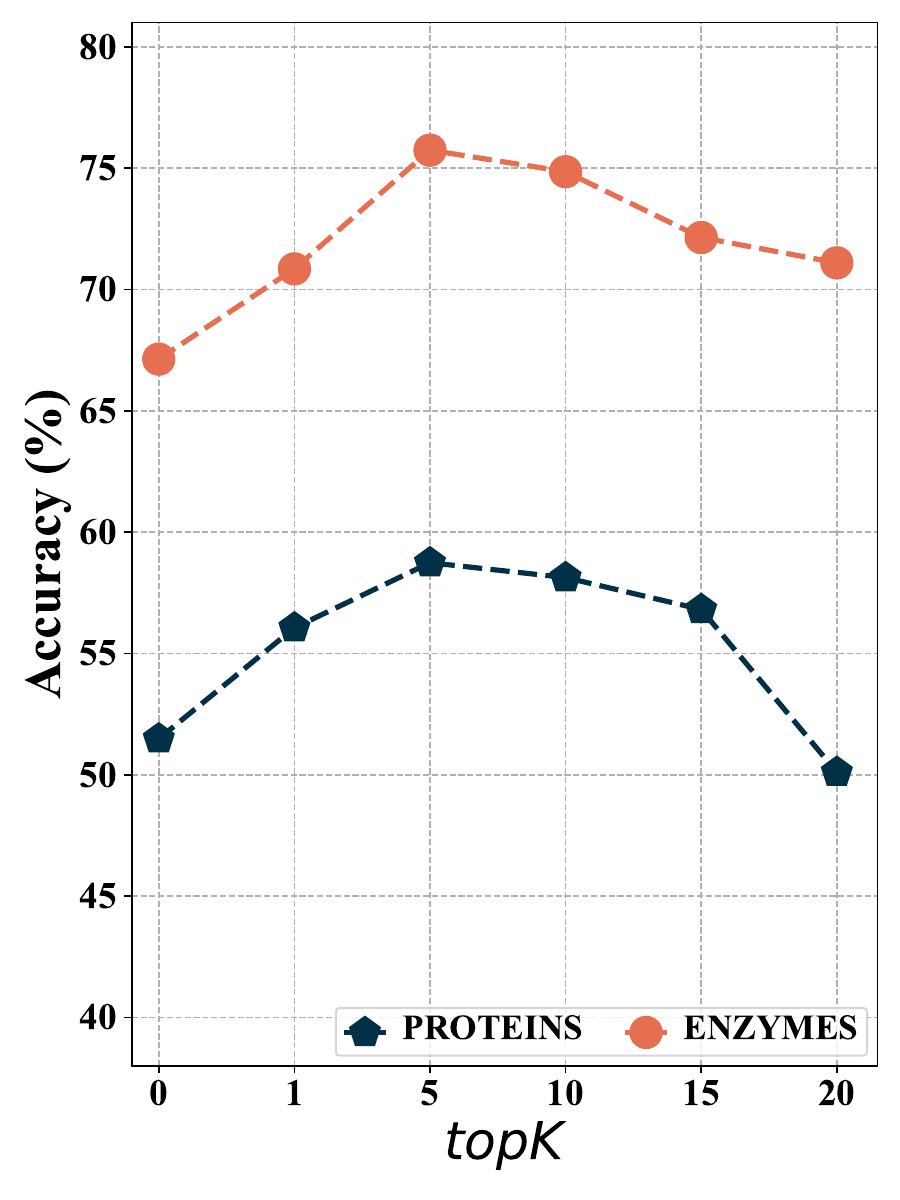}
\vspace{-0.3cm}
\caption{\footnotesize{Hyper-parameter study with hops$k$ (\textbf{Left}) from 1 to 5 and$topk$ from 1 to 20 (\textbf{Right}) on node classification with PROTEINS, and ENZYMES datasets with the setting in Table~\ref{table.combined-classification}.}}
\vspace{-0.3cm}
\label{fig:hyper_study}
\end{wrapfigure}

We observe that PRODIGY/NF and \methodname/NF are better to Vanilla/NF, indicating that the retrieval knowledge truly works when testing on unseen datasets. Moreover, the difference between PRODIGY/NF and PRODIGY/FT is much greater than that of \methodname, which also indicates that a simple learning paradigm for ICL is not enough and that \methodname can achieve acceptable results even on unseen downstream datasets without the need for sophisticated fine-tuning.

\begin{table}[tbp]
\centering\setlength\tabcolsep{1.8pt} 
    \renewcommand{\arraystretch}{1.1} 
    \caption{Accuracy evaluation on node and graph classification. All tabular results (\%) are in mean$\pm$ standard deviation across five seeds run, with best \textbf{bolded} and runner-up \underline{underlined}.}
    \label{table.combined-classification}
    \resizebox{\linewidth}{!}
{
    \begin{tabular}{lcccccc}
    \toprule
    \multirow{3}*{Methods} & \multicolumn{2}{c}{Node Classification} & \multicolumn{4}{c}{Graph Classification} \\
    \cmidrule(lr){2-3} \cmidrule(lr){4-7}
     & PROTEINS & ENZYMES & PROTEINS & COX2 & ENZYMES & BZR \\
    & (5-shot) & (5-shot) & (5-shot) & (5-shot) & (5-shot) & (5-shot) \\
    \midrule\midrule
    GCN & 46.63$\scriptstyle{\pm 03.04}$ & 52.80$\scriptstyle{\pm 12.89}$ & 54.80$\scriptstyle{\pm 06.64}$ & 67.87$\scriptstyle{\pm 03.39}$ & 22.67$\scriptstyle{\pm 05.20}$ & 58.76$\scriptstyle{\pm 05.08}$ \\
    GraphSAGE &48.87$\scriptstyle{\pm 02.64}$ & 48.75$\scriptstyle{\pm 01.59}$ & 52.99$\scriptstyle{\pm 10.57}$ & 67.02$\scriptstyle{\pm 05.42}$ & 21.17$\scriptstyle{\pm 05.49}$ & 58.27$\scriptstyle{\pm 04.79}$ \\
    GAT & 48.13$\scriptstyle{\pm 07.90}$ & 47.75$\scriptstyle{\pm 01.23}$ & 55.82$\scriptstyle{\pm 07.31}$ & 64.89$\scriptstyle{\pm 03.23}$ & 20.67$\scriptstyle{\pm 03.27}$ & 57.04$\scriptstyle{\pm 06.70}$ \\
    GIN & 49.61$\scriptstyle{\pm 01.58}$ & 48.82$\scriptstyle{\pm 01.58}$ & 56.17$\scriptstyle{\pm 08.58}$ & 62.77$\scriptstyle{\pm 02.85}$ & 19.00$\scriptstyle{\pm 03.74}$ & 56.54$\scriptstyle{\pm 04.20}$ \\
    \cmidrule(lr){1-7}
    {\textbf{GraphPrompt}+} \\  \cmidrule(lr){1-7}
    Vanilla/NF & 44.88$\scriptstyle{\pm 13.17}$ & 48.81$\scriptstyle{\pm 01.88}$ & 56.68$\scriptstyle{\pm 03.63}$ & 53.04$\scriptstyle{\pm 04.13}$ & 36.50$\scriptstyle{\pm 03.31}$ & 68.77$\scriptstyle{\pm 03.44}$\\
    Vanilla/FT & 48.99$\scriptstyle{\pm 01.88}$ & 51.99$\scriptstyle{\pm 01.36}$ & 57.04$\scriptstyle{\pm 03.88}$ & 64.04$\scriptstyle{\pm 08.20}$ & 40.00$\scriptstyle{\pm 04.36}$ & 69.01$\scriptstyle{\pm 02.21}$ \\  
    PRODIGY/NF &47.32$\scriptstyle{\pm 08.12}$&43.80$\scriptstyle{\pm14.03}$&53.48$\scriptstyle{\pm06.72}$&53.97$\scriptstyle{\pm10.34}$&22.12$\scriptstyle{\pm13.84}$&67.18$\scriptstyle{\pm08.93}$\\
    PRODIGY/FT &53.26$\scriptstyle{\pm06.42}$&57.98$\scriptstyle{\pm12.37}$&57.14$\scriptstyle{\pm10.34}$&65.31$\scriptstyle{\pm04.28}$&25.94$\scriptstyle{\pm05.12}$&68.08$\scriptstyle{\pm06.68}$\\ \cmidrule(lr){1-7}
    \methodname/NF & {56.12}$\scriptstyle{\pm 04.11}$ & \underline{75.92}$\scriptstyle{\pm 01.72}$ & {58.48}$\scriptstyle{\pm 03.93}$ & 55.32$\scriptstyle{\pm 04.15}$ & 38.17$\scriptstyle{\pm 03.39}$ & \textbf{77.53}$\scriptstyle{\pm 05.26}$\\
    \methodname/FT & \underline{58.74}$\scriptstyle{\pm 00.87}$ & {75.74}$\scriptstyle{\pm 01.92}$ & \textbf{62.33}$\scriptstyle{\pm 02.52}$ &\textbf{76.60}$\scriptstyle{\pm 02.30}$ &\underline{47.71}$\scriptstyle{\pm 06.88}$ & \underline{76.79}$\scriptstyle{\pm 05.02}$\\
    \methodname/NFT & \textbf{59.83}$\scriptstyle{\pm 00.40}$ & \textbf{76.23}$\scriptstyle{\pm 01.63}$ & \underline{59.08}$\scriptstyle{\pm 03.50}$ & \underline{71.70}$\scriptstyle{\pm 04.29}$ & \textbf{49.17}$\scriptstyle{\pm 04.64}$ & 74.81$\scriptstyle{\pm 04.25}$\\
    \bottomrule
    \end{tabular}}
\end{table}

\begin{table}[tbp]
    \centering
    \renewcommand{\arraystretch}{1.1} 
    \caption{Performance evaluation (‰) on link prediction.}
    \label{tab:overall}
\setlength\tabcolsep{1.7pt} 
\resizebox{\linewidth}{!}
{
    \begin{tabular}{lcccccc}
        \toprule
        \multirow{2}{*}{Method} & \multicolumn{2}{c}{TAOBAO} & \multicolumn{2}{c}{KOUBEI} & \multicolumn{2}{c}{AMAZON} \\
        \cmidrule(lr){2-3}\cmidrule(lr){4-5}\cmidrule(lr){6-7}
        ~ & Recall & nDCG & Recall & nDCG & Recall & nDCG \\
        \midrule
        LightGCN & 22.47$\scriptstyle{\pm 02.53}$ & 21.89$\scriptstyle{\pm 02.80}$ & 30.21$\scriptstyle{\pm 06.45}$ & 22.24$\scriptstyle{\pm 05.83}$ & 15.07$\scriptstyle{\pm 06.48}$ & 06.53$\scriptstyle{\pm 02.66}$ \\ 
        SGL & 22.15$\scriptstyle{\pm 02.20}$ & 22.12$\scriptstyle{\pm 03.09}$ & 32.61$\scriptstyle{\pm 04.27}$ & 22.36$\scriptstyle{\pm 04.82}$ & 15.78$\scriptstyle{\pm 07.12}$ & 07.90$\scriptstyle{\pm 02.49}$ \\ 
        MixGCF & 22.84$\scriptstyle{\pm 02.15}$ & 23.05$\scriptstyle{\pm 03.87}$ & 32.06$\scriptstyle{\pm 04.20}$ & 22.49$\scriptstyle{\pm 06.91}$ & 15.24$\scriptstyle{\pm 08.98}$ & 07.40$\scriptstyle{\pm 03.44}$ \\ 
        SimGCL & 22.18$\scriptstyle{\pm 02.22}$ & 23.15$\scriptstyle{\pm 02.75}$ & 33.07$\scriptstyle{\pm 05.28}$ & 23.08$\scriptstyle{\pm 05.55}$ & 16.10$\scriptstyle{\pm 07.91}$ & 07.58$\scriptstyle{\pm 03.51}$ \\
        \cmidrule(lr){1-7}
        \textbf{GraphPro}+ \\
        \cmidrule(lr){1-7}
        Vanilla/NF  & 20.10$\scriptstyle{\pm 01.50}$ & 20.12$\scriptstyle{\pm 01.30}$ & 21.31$\scriptstyle{\pm 04.59}$ & 15.31$\scriptstyle{\pm 03.11}$ & 12.56$\scriptstyle{\pm 07.45}$ & 06.31$\scriptstyle{\pm 03.92}$ \\
        Vanilla/FT  & {23.99}$\scriptstyle{\pm 02.11}$ & 23.26$\scriptstyle{\pm 01.42}$ & \underline{33.96}$\scriptstyle{\pm 04.13}$ & \underline{24.66}$\scriptstyle{\pm 02.78}$ & 18.14$\scriptstyle{\pm 07.55}$ & 08.73$\scriptstyle{\pm 03.74}$ \\
        PRODIGY/NF  & 21.67$\scriptstyle{\pm 01.42}$ & 23.15$\scriptstyle{\pm 03.20}$ & 21.66$\scriptstyle{\pm 03.21}$ & 14.82$\scriptstyle{\pm 03.92}$ & 11.88$\scriptstyle{\pm 02.61}$ & 05.84$\scriptstyle{\pm 01.84}$ \\
        PRODIGY/FT  & {23.74}$\scriptstyle{\pm 01.22}$ & 23.65$\scriptstyle{\pm 02.31}$ &{33.46}$\scriptstyle{\pm 04.70}$ &{23.28}$\scriptstyle{\pm 03.40}$ & 16.72$\scriptstyle{\pm 04.28}$ &{08.09}$\scriptstyle{\pm 02.66}$ \\
        \cmidrule(lr){1-7}
        \methodname/NF  & 20.31$\scriptstyle{\pm 01.60}$ & 20.45$\scriptstyle{\pm 01.44}$ & 22.86$\scriptstyle{\pm 03.44}$ & 16.68$\scriptstyle{\pm 02.48}$ & 13.78$\scriptstyle{\pm 05.54}$ & 06.52$\scriptstyle{\pm 02.69}$ \\
        \methodname/FT  &\underline{24.78}$\scriptstyle{\pm 01.93}$ &\underline{24.35}$\scriptstyle{\pm 01.34}$ & \textbf{34.27}$\scriptstyle{\pm 03.93}$ & \textbf{24.82}$\scriptstyle{\pm 02.69}$ &\textbf{18.69}$\scriptstyle{\pm 07.45}$ & \textbf{09.09}$\scriptstyle{\pm 03.89}$ \\
        \methodname/NFT  &\textbf{24.89}$\scriptstyle{\pm 02.01}$ &\textbf{24.51}$\scriptstyle{\pm 01.44}$ &{33.27}$\scriptstyle{\pm 04.37}$ &{24.09}$\scriptstyle{\pm 03.14}$ &\underline{18.32}$\scriptstyle{\pm 06.91}$ &\underline{09.01}$\scriptstyle{\pm 03.67}$ \\ 
        \bottomrule
    \end{tabular}
}
\end{table}
\subsection{Hyper-parameter Study}
In this section, we examine the impact of various hyper-parameters on \methodname. We specifically analyze the effects of varying the number of hops \( k \) in toy graphs from the list $[1, 2, 3, 4, 5]$ 
and the number of linked toy graphs \( topK \) from the list \([1, 5, 10, 15, 30, 50]\) to verify the sensitive: 

Figure~\ref{fig:hyper_study} (Left) illustrates relationships between accuracy and the toy graph hop \( k \). We observe that as \( k \) increases, the volume of retrieved knowledge grows exponentially. However, an excessive accumulation of knowledge not only fails to enhance accuracy but also introduces increased irrelevant noise that burdens the GNNs. Notably, accuracy shows a trend of initial improvement followed by a decline as \( k \) is increased. This pattern suggests that at lower \( k \) values, the retrieved information tends to consist of isolated, less useful knowledge. In contrast, at higher \( k \) values, the GNNs struggle to process extensive reasoning chains, leading to the utilization of complex and abundant information that is less effective than even the baseline model's performance.
Figure~\ref{fig:hyper_study} (Right) shows effects on accuracy with different numbers of toy graphs \( topK \). As with the previous figure, increasing \( topK \) demonstrates that an excessive amount of knowledge can hinder the GNNs' comprehension capabilities. Conversely, smaller \( topK \) results in insufficient knowledge to enhance performance on downstream tasks.
\section{Conclusion}
We introduced \methodname, a novel and general framework that enhances Graph Neural Networks (GNNs) by integrating Retrieval-Augmented Generation (RAG) techniques. This plug-and-play approach improves GNNs' ability to generalize to unseen data by retrieving relevant information. Experimental results show that \methodname outperforms state-of-the-art methods in various graph learning tasks, demonstrating its adaptability and robustness. While \methodname is currently limited to retrieving subgraphs, future research could explore using more graph-structured data such as nodes, edges, and trees to further enhance its capabilities. In general, our work provides valuable insights and serves as a reference for future Large Graph Models.

\section*{Acknowledgments}
This work is supported by the National Natural Science Foundation of China (No.U23A20468).

\newpage
\bibliography{example_paper}
\bibliographystyle{abbrv}

\newpage
\appendix
\onecolumn
\section{Notations}
\label{sec:app notation}
The notations in this paper are summarized in Table~\ref{tab:symbols}.
\begin{table}[htb]
    \centering
      \caption{Notations Tables in \methodname}
    \label{tab:symbols}
    \begin{tabular}{c|l}        \hline
        \rowcolor[gray]{0.94} Notation & Definition 
        \\ \hline
        $\mathcal{G}$ / $\mathcal{V}$ / $\mathcal{E}$ & The dynamic graph / node / edge set with $G_t$/ $V_t$ / $E_t$ as its entry \\
        $\mathcal{G}^\mathcal{R}$ & The resource graph with $G^\mathcal{R}_t$ as its entry\\ 

        ${G}^\mathcal{Q}$ & The query graph with $v_c$ as its center node \\
        
        $\mathcal{G^T}$ & The toy graph database \\ 
        $G^\mathcal{T}_{\text{TopK}}$ & The topK retrieved toy graphs \\ 
        ${G}^\mathcal{T}$ & The toy graph \\ $G^{e}_\tau(v_m)$ &
        $k$-hop ego net for node $v_m$
        \\
        $X_t$ / $A_t$ / $Y_t$ & The feature / edge weight / label matrix at time $t$\\
        $\mathcal{C}$ & The set of label classes \\
        $\mathcal{N}(v)$ & The neighbors of node $v$ \\ 
        $H$ / $O$ & The hidden embeddings / task-specific output vector with $h_i$ / $o_i$ as its $v_i$ vector\\ 
        $\mathcal{D}$ & The $\kappa$-shot labeled set \\ 
        \hdashline
        $t / \tau$ & Timestamp \\
        $n$ & Number of nodes \\
        $d$ & The dimension of node feature \\
        $v_i$ & The $i$-th node \\
        $v_m / v_c$ & The master node of toy graph / The center node of toy graph \\
        $f_1 / f_2$ & The dimension of hidden embedding / task-specific output vector \\ 
        $\epsilon$ & The minimum value \\
        $k$ & $k$ hop \\
        \hdashline
        $I(v)$ & The node importance for node $v\in G^\mathcal{R}$ \\
        $\text{PR}(\cdot) / \text{DC}(\cdot)$ & The PageRank / Degree Centrality value \\
        $p_i$ & The sampling probability of node $v_i$ \\
        $K$ & The scaling constant\\
        $n_{\text{aug}}(G^{\mathcal{T}})$ & The number of augmentations of toy graph $G^{\mathcal{T}}$ \\
        \hdashline 
        $S(v_c,v_m)$ & The weighted similarity between query node $v_c$ and master  node $v_m$ \\
        $l$ & The layer of a GNN \\
        $\alpha$ & The balance weight with $\alpha \in (0,1)$\\
        $\lambda$ & The weight of mixup \\
        $w_i$ &  The weights of the time, structure, semantic, and environment similarities\\ 
        $\gamma$ & The reweight hyper-parameter \\
         \hline
    \end{tabular}
\end{table}

\section{More Motivation Details}
\subsection{Why Toy Graph Augmentation is needed}
The reasons for toy graph augmentation:
\begin{itemize}[leftmargin=*]
    \item Expanding toy graph base, enriching the scale of the knowledge repository~\cite{9156551}.
    \item Simulating Real-World Scenarios: Real-world graphs often encounter challenges such as missing nodes~\cite{111111}, noisy attributes~\cite{Li2024BoostingTA}, and unexplored connections~\cite{10146293}. We introduce node dropout, noise injection, and edge removal to simulate these scenarios accurately.
    \item Addressing Graph Domain Shift: To mitigate domain shift between the graph knowledge base and testing graphs, our augmentations employ Mixup techniques such as Node Interpolation and Edge Rewiring. These techniques interpolate between training samples to generate synthetic samples, effectively smoothing decision boundaries in embedding and reducing the model's sensitivity to minor variations in input data, thereby stabilizing predictions on domain shift testing samples~\cite{Xu2024ProtoMixAH}.

\end{itemize}

\subsection{Why Noise-based Graph Prompt Tuning is needed}
To address inherent challenges in toy graph quality, we introduce Noise-based Graph Prompting Tuning (c.f. Section 4.3.3). This method involves fine-tuning the model with artificially introduced noisy toy graphs (Inner-Toy-Graph Noise \& Toy-Graph Noise), inspired by noise-tuning techniques in NLP~\cite{fang2024enhancingnoiserobustnessretrievalaugmented,Cuconasu_2024,yoran2023making}. Our approach enhances the model's robustness against real-world retrieval noise, as evidenced by superior performance compared to traditional tuning methods (in Main Text Tables 1 and 2). This approach reduces the stringent requirement for an exceptionally high-quality graph vector base, thereby ensuring robust performance across various tasks within our \methodname, and significantly mitigating data quality impacts.

\subsection{Difficulty to construct and maintain high-quality and diverse graph vector base}
In \methodname, the toy graph base largely leverages significant prior research datasets in pre-trained GNNs~\cite{graphprompt,xia2024opengraph,mishchenko2023prodigy,yu2023multigprompt}, which are trained on meticulously curated graph datasets and cover diverse domains, such as biology, chemistry, medicine recommendation tasks, etc. For example, the PROTEINS dataset~\cite{borgwardt2005protein}, derived from cryo-electron microscopy and X-ray crystallography, and the ENZYMES dataset~\cite{wang2022faith}, based on EC enzyme classification, are meticulously annotated by medical experts.

\subsection{Why Inverse Importance Sampling Strategy is needed}
The adoption of the Inverse Importance Sampling strategy is crucial. In \methodname, subgraphs are sampled as toy graphs, where nodes with higher degrees (non-long-tail knowledge, extensively learned and embedded into GNN parameters) are more frequently included in subgraphs due to their extensive connections with neighbors, resulting in higher frequency in toy graph base~\cite{5462078}. Conversely, nodes with low degrees (long-tail knowledge), are more important but ignored. To mitigate this issue, we propose this by prioritizing nodes with lower degrees to capture long-tail knowledge when sampling.

\subsection{Why Four Similarities are needed}
In practical applications, the four similarities all contribute to performance improvement and we state the significance as follows:
\begin{itemize}[leftmargin=*]
    \item Time information is crucial to predict future states or trends~\cite{yang2024graphpro} via node history, i.e. in social networks, analyzing historical user interaction aids in predicting future behaviors.
    \item Structure pertains to how nodes are interconnected and overall graph topology, vital for capturing similar graph structure patterns~\cite{cui2021positional,111111,yang2024faima}. In transportation networks, factories are always located on the outer ring of the city, sharing similar structural connectivity, aiding in the discovery of spatiotemporal patterns~\cite{SPGCL,Duan2024CausalDA}.
    \item Sharing similar neighborhoods is essential for evaluating node similarity and correlation. In recommendations, shared purchase histories between users and products indicate potential interests, akin to collaborative filtering~\cite{10421425}.
    \item Semantic information measures similarity based on features~\cite{mishchenko2023prodigy}. In knowledge graphs, identifying relevant subgraphs to query nodes enhances retrieval accuracy based on semantic similarity.
\end{itemize}

\subsection{Why Knowledge Fusion is needed}
Fusion and decoder here represent one of the core contributions of \methodname:
\begin{itemize}[leftmargin=*]
    \item Overall Task Perspective: For the same tasks, the decoder can be directly employed to obtain outputs. For different tasks, the decoder can be masked and utilize pre-computed embeddings without training or be tuned to better adapt. This underscores our primary contribution, where the decoder functions as a versatile "plug-and-play" and "tune-free" component.
    \item Integral Fusion Strategy: Fusion Strategy facilitates concurrent information propagation from toy graphs X (hidden embeddings) and Y (task-specific output vector) to query graph, aligning with our secondary contribution.
\end{itemize}

\subsection{How \methodname works on par with RAG in NLP and CV}
In NLP, RAG enhances the generation of LLM by retrieving relevant information via prompts. Similarly, in \methodname, we enhance downstream graph learning by integrating information from retrieved toy graphs. Using these toy graphs with shared patterns assists the model inference. In our framework, the "generation" involves the retrieval-enhanced Graph Prompt: Toy Graph Intra Propagate \& Query-Toy-Graph Inter Propagate to propagate retrieved knowledge (X and Y) into the query graph. To illustrate, we analyze this from both experiment and theory.

\begin{enumerate}[leftmargin=*]
    \item \textbf{Experiment 1:} We perform a case study to illustrate how "generation" works by displaying specific instances of node vectors in Appendix~\ref{app:works}.

    \item \textbf{Experiment 2:} In traditional GNN tasks, GCN, GAT, and GIN typically expand their receptive fields through stacked message-passing layers or neighborhood subgraph sampling for inference. Patterns learned in these contexts are often localized within the constrained receptive field. In contrast, in \methodname, we observe that subgraphs sharing similar patterns often exhibit properties more aligned with downstream tasks. These subgraphs provide richer information for inference compared to simply enlarging receptive fields. As shown in Main Text Tables 1 and 2, Figure 3, \methodname's strategy of incorporating toy graphs significantly outperforms baselines.

    \item \textbf{Theory 1:} Furthermore, we provide a theoretical justification of retrieval augmentation in GNNs (see Appendix B.4). From an information-theoretic perspective, introducing RAG knowledge into GNNs enhances the mutual information between input features $X$ and output labels $Y$, such that:
    \[
    I(X, RAG; Y) \geq I(X; Y),
    \]
    thereby improving the performance of downstream tasks. This is aligned with the information theory of RAG in NLP~\cite{NEURIPS2022_1eeaae7c}.

    \item \textbf{Theory 2:} Recent studies~\cite{108652,zhunformation} also suggest the generalization error diminishes with an increase in the node number of the graph in Theorem 1.1~\cite{zhunformation}: the generalization error between the expected loss $R_{\text{exp}}(\Theta) = \mathbb{E}_{(x,y) \sim \mu_G}[\mathcal{L}(\Theta(x), y)]$ and empirical loss $R_{\text{emp}}(\Theta) = \frac{1}{m}\sum_{i=1}^{m} \mathcal{L}(\Theta(x^i), y^i)$ are super bounded:
    \[
    |R_{\text{exp}}(\Theta) - R_{\text{emp}}(\Theta)| \leq \sqrt{\frac{C}{m} q(n)},
    \]
    where $C$ represents the model complexity (e.g., parameters), $m$ denotes the training set size, and $q(n) = \mathbb{E}_{n \sim \nu_G}[n^{-\frac{1}{D+1}}]$ depends on the average graph size (node number) with $\nu$ as the graph size distribution and $D$ is the metric-measure space dimension. In \methodname, retrieving similar toy graphs significantly increases the number of graph nodes (via Query-Toy-Graph Inter Propagate, linking toy graph nodes to query graph), significantly augmenting $n$ while reducing $q(n)$. Consequently, the upper bound of generalization error decreases, promoting smoother graph learning convergence and enhancing pattern learning.
\end{enumerate}

\section{Further Methods Details}
\subsection{Revisiting Graph Neural Networks}
\label{sec: app revisiting GNN}
{The goal of a GNN is to learn node embeddings based on an iterative aggregation of messages from the local network neighborhood. We use embedding matrix $\{\mathbf{z}_v^{(L)}\}_{v \in \mathcal{V}}$ to denote the embedding for all the nodes after applying an $L$-layer GNN. The $l$-th layer of a GNN, $\{\mathbf{z}_v^{(L)}\} = \textsc{GNN}^{(l)}(\{\mathbf{z}_v^{(l-1)}\})$, can be written as:
\begin{equation}
    \begin{aligned}
    &\mathbf{m}_{u \to v}^{(l)} = \textsc{Msg}^{(l)}(\mathbf{z}_u^{(l-1)}, \mathbf{z}_{v}^{(l-1)}), \\
    &\mathbf{z}_v^{(l)} = \textsc{Agg}^{(l)}\big(\{\mathbf{m}_{u \to v}^{(l)} \mid u \in \mathcal{N}(v)\}, \mathbf{z}_v^{(l-1)}\big),
    \end{aligned}
    \label{eq:gnn}
\end{equation}
where $\mathbf{z}_v^{(l)}$ is the embedding for $v \in {V}$ after passing through $l$ layers, $\mathbf{z}_v^{(0)}={x}_v ~\text{or}~ {h}_v ~\text{or}~ {o}_v$, $\mathbf{m}_{u \to v}^{(l)}$ is the message embedding, and $\mathcal{N}(v)$ is the set of direct neighbors of $v$. Different GNNs can have various definitions of message-passing functions $\textsc{Msg}^{(l)}(\cdot)$ and aggregation functions $\textsc{Agg}^{(l)}(\cdot)$ and these two functions could be parameter-free.}

\subsection{Key Construction of Position-aware Code}
\label{sec:key construction}
Given a randomly selected node anchor set $ \mathcal{V_S} \subset \mathcal{V}$, we calculate the minimal distances, \emph{a.k.a.} hops between the two node sets. Suppose $v_u \in \mathcal{V}, v_w \in \mathcal{V_S}$, the distance similarity between node $v_u$ and $v_w$ can be depicted as $dis(v_u, v_w)$. By normalizing the similarity to [0, 1], distance-to-centroid $d2c(v_u,v_w)$: 
\begin{equation}
        d2c(v_u,v_w)=\left\{
        \begin{alignedat}{3}
        &\frac{1}{dis(v_u, v_w)+1}, &\quad&\text{if }  dis(v_u, v_w)<dis_q \\
        &0, &\quad&\text{otherwise}
\end{alignedat},
\right.
    \label{eq:d2c}
\end{equation}
here hyperparameter $dis_q$ is the maximum hops, the distance beyond this boundary is considered invalid.
The structure feature of node $v_u$ is $d2c(v_u, \mathcal{V_S})$. By collecting all distances with anchor-set $\mathcal{V_S}$, the structure $S\in R ^{n\times |\mathcal{V_S}|}$ is written as follows:
\begin{equation}
 \begin{aligned}
&d2c(v_u, \mathcal{V_S})= [d2c(v_u, v_w)|v_w\in \mathcal{V_S}\subset \mathcal{V}],
\\ & S = [d2c(v_u, \mathcal{V_S})|v_u\in \mathcal{V}],
    \label{eq:build S}
 \end{aligned}
\end{equation}
where $[\cdot]$ means the concatenation operation.
The distance-to-centroid feature converts the non-Euclidean structure to the Euclidean structure. $d2c$ dramatically reduces the size of the matrix and meanwhile contains more structure information instead of identifier information, the size of the anchor set is $\log_2 n $ follows P-GNNs\cite{P-GNN,111111}.

\subsection{Similarity Functions}
\label{sec:app sim func}
For the {history key}, we adopt an exponential decay function to measure the time similarity values. We smooth the impact of time differences and provide a controlled decay coefficient $\eta> 0$. The time similarity, \( S_\text{time} \), between the same node \( v_c \) and \( v_m \) with different timestamp \( t(v_m),t(v_c) \), is defined as:
\begin{equation}
S_\text{time}(v_c, v_m) = e^{-\eta |t(v_c) - t(v_m)|},
\end{equation}
where \( \eta \) is a positive parameter that controls the rate of exponential decay. 

For the environment key, we match the environment of node \( v \) using Jaccard similarity to compare the sets of neighbors \( \mathcal{N}{(v_c)} \) in the query graph and \( \mathcal{N}{(v_m)} \) in the toy graph:
\begin{equation}
S_\text{environment}(v_c, v_m) = \frac{|\mathcal{N}{(v_c)} \cap \mathcal{N}{(v_m)}|}{|\mathcal{N}{(v_c)} \cup \mathcal{N}{(v_m)}|}.
\end{equation}

For the hidden embedding key, we input the query graph into pre-trained GNNs to obtain the hidden embedding for the query node, with the similarity defined as:
\begin{equation}
h_c = \textsc{Gnn}(X_{G^{\mathcal{Q}}}), \quad S_\texttt{semantic}(v_c, v_m) = \texttt{cosine}(h_c, h_m).
\end{equation}

For the position-aware code, we denote $s_c,s_m$ as the position-aware code of node $v_c, v_m$, and utilize cosine similarity as before, defined as \( S_\text{structure}(v_c, v_m) = \texttt{cosine}(s_c, s_m) \).

\subsection{Proof of the Effectiveness of RAG}
\label{sec: app effectiveness of rag}
In this section, we will theoretically prove that introducing RAG knowledge can significantly improve the predictive performance of the model. 

Assume \(X\) represents the input features, \(Y\) represents the target output labels, and \(RAG\) represents external knowledge related to the input features (or even the output labels). 
We analyze from the mutual information view, where \(I(X; Y)\) quantifies the dependency between \(X\) and \(Y\), which reflects the performance of the model, the larger the value, the better the performance of the model~\cite{SPGCL,Mao_2021}. 
By introducing RAG knowledge $RAG$ into GNNs, we can effectively increase the mutual information between the input features \(X\) and the output labels \(Y\) as $I(X, RAG; Y)\ge I(X; Y)$, thereby improve the model's downstream task performance.
The derivation is as follows:
\begin{align}
    & I(X, RAG; \mathbf{Y}) - I(X; \mathbf{Y}) \nonumber \\
    =& \sum_{X, RAG, \mathbf{Y}} p(X, RAG, \mathbf{Y}) \log \frac{p(X, RAG, \mathbf{Y})}{p(X, RAG) p(\mathbf{Y})} - \sum_{X, \mathbf{Y}} p(X, \mathbf{Y}) \log \frac{p(X, \mathbf{Y})}{p(X) p(\mathbf{Y})} \nonumber \\
    =& \sum_{X, RAG, \mathbf{Y}} p(X, RAG, \mathbf{Y}) \log \frac{p(X, RAG, \mathbf{Y})}{p(X, RAG) p(\mathbf{Y})} - \sum_{X, RAG, \mathbf{Y}} p(X, RAG, \mathbf{Y}) \log \frac{p(X, \mathbf{Y})}{p(X) p(\mathbf{Y})} \nonumber \\
    =& \sum_{X, RAG, \mathbf{Y}} p(X, RAG, \mathbf{Y}) \log \left( \frac{p(X, RAG, \mathbf{Y})}{p(X, RAG)} \cdot \frac{p(X)}{p(X, \mathbf{Y})} \right) \nonumber \\
    =& \sum_{X, RAG, \mathbf{Y}} p(X, RAG, \mathbf{Y}) \log \frac{p(X, RAG, \mathbf{Y})}{p(RAG \mid X) p(X, \mathbf{Y})} \nonumber \\
    =& \sum_{X, RAG, \mathbf{Y}} p(RAG, \mathbf{Y} \mid X) p(X) \log \frac{p(RAG, \mathbf{Y} \mid X)}{p(RAG \mid X) p(\mathbf{Y} \mid X)} \nonumber \\
    =& I(RAG; \mathbf{Y} \mid X) \geq 0,
\end{align}
where \( \sum_{X, RAG, \mathbf{Y}} = \sum_{X} \sum_{RAG} \sum_{\mathbf{Y}}\).
Moreover, $I(RAG; \mathbf{Y} \mid X)$ measures that how much additional knowledge $RAG$ provides to assist in predicting $Y$ based on $X$, this term will approach zero when the $RAG$ is noise to the prediction task.
In summary, the integration of RAG knowledge can enhance the mutual information between \(X\) and \(Y\), thereby improving the performance and accuracy of downstream tasks.

\subsection{Algorithms}
\label{app:sec alg}
In this section, we will provide a detailed description of the algorithms of Toy Graph Construction (Algorithm~\ref{alg.toygraph}) and Training and Inference with Toy Graphs Retrieval (Algorithm~\ref{alg.model}).

\begin{algorithm}[h]
\caption{Toy Graph Construction}
\label{alg.toygraph}
\begin{algorithmic}[1]
    \Require Dynamic Resource Graph $\mathcal{G^R}$, node importance balance weight $\alpha$, toy graph scaling constant $K$, maximum hop $k$
    \Ensure Toy graph embedding key-value database $\mathcal{G}^\mathcal{T}$
    
    \State Initialize toy graph database $\mathcal{G}^\mathcal{T} \leftarrow \emptyset$
    \For{each snapshot $G_\tau^\mathcal{R} \in \mathcal{G^R}$} \Comment{\textbf{Construct Toy Graphs}}
        \For{each node $v \in G_\tau^\mathcal{R}$}
            \State Calculate importance $I(v) \leftarrow \alpha \textsc{PR}(v) + (1 - \alpha) \textsc{DC}(v)$
            \State Reverse node importance $I'(v) \leftarrow \frac{1}{I(v) + \epsilon}$
        \EndFor
        \For{each node $v \in G_\tau^{\mathcal{R}}$}
            \State Normalize sampling probabilities $p_i \leftarrow \frac{I'(v_i)}{\sum_{j=1}^n I'(v_j)}$
        \EndFor
        \State Sample master node $v_m \leftarrow$ \textsc{WeightedSampling}($G_\tau^{\mathcal{R}}$, $p_i$) based on probability $p_i$
        \State Generate $k$-hop ego net $G^{e}_\tau(v_m)$ for node $v_m$
        \State Augment toy graph $\{G^{\mathcal{T}}\} \leftarrow$ \textsc{DataAugmentation}($G^{e}_\tau(v_m)$, $n_{\text{aug}})$ with $n_{\text{aug}}(G^{e}_\tau(v_m)) = \lfloor K \cdot \bar{I'}(G^{e}_\tau(v_m)) \rfloor$
        \For{each $G^{\mathcal{T}} \in \{G^{\mathcal{T}}\}$} \Comment{\textbf{Generate keys-values pair}}
            \State Save timestamp $\tau$ as history key
            \State Save neighbors $\mathcal{N}(v_m^\tau)$ of master node $v_m$ as environment key
            \State Save structural encoding $s_m^\tau$ of node $v_m$ via Eq.~(\ref{eq:build S}) as structure key
            \State Save hidden embedding $h_m^\tau$ by feeding $G^{\mathcal{T}}$ into pre-trained GNNs as semantic key
            \State Save the hidden embedding $\{h_i^\tau | v_i \in G^{\mathcal{T}}\}$ as value
            \State Save task-specific output vectors $\{o_i^\tau | v_i \in G^{\mathcal{T}}\}$ by feeding $\{h_i^\tau | v_i \in G^{\mathcal{T}}\}$ into decoder as value
            \State Store toy graph $G^{\mathcal{T}}$ into database $\mathcal{G}^\mathcal{T}$
        \EndFor
    \EndFor
    \State \Return Toy graph database $\mathcal{G}^\mathcal{T}$
\end{algorithmic}
\end{algorithm}

\paragraph{\textbf{Algorithm 1.} }{We outline the process for constructing toy graphs in Algorithm.~\ref{alg.toygraph}. In line 1, the toy graph database $\mathcal{G}^\mathcal{T}$ is initialized. 

Lines 2-15 describe the steps to construct toy graphs by iterating through each snapshot of the dynamic resource graph $\mathcal{G^R}$.

In more detail, lines 3-5 details calculate the importance and reverse importance of each node within the snapshot. Following this, lines 6-7 involve normalizing the sampling probabilities according to the reverse importance values. The selection of a master node and the generation of its $k$-hop ego network are carried out in lines 8-9. Subsequently, line 10 involves augmenting the toy graph through specific data augmentation techniques.

Lines 11-16 detail the generation of key-value pairs for each toy graph. This includes saving the timestamp as the history key, the neighbors of the master node as the environment key, the structural encoding as the structure key, the hidden embedding as the semantic key, and the task-specific output vectors as the value. Each toy graph is then stored in the database $\mathcal{G}^\mathcal{T}$.

Ultimately, in line 18, the algorithm returns the toy graph database $\mathcal{G}^\mathcal{T}$.}

\paragraph{\textbf{Algorithm 2.} }
We introduce the algorithm for training and inference with toy graph retrieval in Algorithm.~\ref{alg.model}. Initially, in line 1, we define the required inputs, including the testing graph $\mathcal{G}_{\text{test}}$, the toy graph database $\mathcal{G}^\mathcal{T}$, and other relevant parameters. The final output is the aggregated result $\tilde{o}_c$.

The \textsc{RetrieveToyGraphs} function, detailed in lines 3-11, initializes an empty similarity list and iterates over each toy graph in the database. Lines 5-6 compute various similarity metrics, and the overall similarity is determined in line 7. This similarity score is then added to the list. After sorting by similarity, the $topK$ toy graphs are retrieved and returned in line 11.

Within the \textsc{Propagation} function (lines 12-17), each retrieved toy graph undergoes intra-propagation in line 14. The intra-propagation step follows in line 16, ultimately returning the propagated results $\mathbf{z}_c$.

The \textsc{KnowledgeFusion} function, found in lines 18-21, combines the outputs from previous steps. The final combined outputs $\tilde{o}_c$ are generated by the decoder and returned in line 21.

The main algorithm begins in line 22. If fine-tuning is enabled and the prompt loss has not converged, lines 23-34 detail the process of toy graph retrieval and propagation for each query graph. This includes the optional addition of noise in lines 26-29. The hidden embedding and task-specific output vectors are propagated in lines 30-31, and the aggregated outputs are fused in line 32. The prompt loss is computed, and fine-tuned parameters are updated in lines 33-34.

If fine-tuning is not required, lines 35-39 describe a similar process of toy graph retrieval and propagation, without the fine-tuning steps. The aggregated outputs are computed directly.

\begin{algorithm}[p]
\caption{Training and Inference with Toy Graphs Retrieval}
\label{alg.model}
\begin{algorithmic}[1]
    \Require Testing graph $\mathcal{G}_{\text{test}}$, toy graph database $\mathcal{G}^\mathcal{T}$, pre-trained GNN model $\text{GNN}_{\Theta_0}(
    \cdot)$, number of $TopK$ toy graphs to retrieve, similarity weights $w_1, w_2, w_3, w_4$, fine-tuning flag $fine\_tune$, noise prompt-tuning flag $add\_noise$
    \Ensure Aggregated output $\tilde{o}_c$

    \Function{RetrieveToyGraphs}{$G^{\mathcal{Q}}$, $\mathcal{G}^\mathcal{T}$, $TopK$}
        \State Initialize similarity list $\{S\} \leftarrow \emptyset$
        \For{each toy graph $G^{\mathcal{T}} \in \mathcal{G}^\mathcal{T}$}
        \State Calculate time similarity $S_\text{time}$, environment similarity $S_\text{environment}$, structure similarity $S_\text{structure}$ and semantic similarity $S_\text{semantic}$
            \State Compute similarity $S \leftarrow w_1 \cdot S_\text{time} + w_2 \cdot S_\text{structure} + w_3 \cdot S_\text{environment} + w_4 \cdot S_\text{semantic}$
            \State Add $(G^{\mathcal{T}}, S)$ to $\{S\}$
        \EndFor
        \State Sort $\{S\}$ by similarity in descending order
        \State Retrieve $topK$ toy graphs $G^\mathcal{T}_{\text{TopK}} \leftarrow \{G^{\mathcal{T}} \in \{S\} \text{ with } topK \text{ similarities}\}$
        \State \Return Retrieved toy graphs $G^\mathcal{T}_{\text{TopK}}$
    \EndFunction

    \Function{Propagation}{$G^{\mathcal{Q}}$, $G^\mathcal{T}_{\text{TopK}}$}
        \For{each toy graph $G^{\mathcal{T}} \in G^\mathcal{T}_{\text{TopK}}$}
            \State Perform Intra Propagation $\mathbf{z}_m \leftarrow$ \textsc{IntraPropagation}($G^{\mathcal{T}}$)
            \EndFor
            \State $\mathbf{z}_c \leftarrow$ \textsc{InterPropagation}($G^{\mathcal{Q}}$, $G^\mathcal{T}_{\text{TopK}}$)
        
        \State \Return $\mathbf{z}_c$
    \EndFunction

    \Function{KnowledgeFusion}{${h}_c,{o}_c$}
            \State Combined output $\hat{o}_c \leftarrow \gamma o_c + (1-\gamma) \textsc{Decoder}(h_c)$
        \State \Return Combined outputs $\tilde{o}_c$
    \EndFunction

    \If{$fine\_tune$ \& $\mathcal{L}_{\text{prompt}}$ not converged} 
        \For{each query graph $G^{\mathcal{Q}} \in \mathcal{G}_{\text{test}}$} \Comment{Toy Graph Retrieval and Propagation}
            \State $G^\mathcal{T}_{\text{TopK}} \leftarrow$ \Call{RetrieveToyGraphs}{$G^{\mathcal{Q}}$, $\mathcal{G}^\mathcal{T}$, $TopK$}
            \If{$add\_noise$}
                \For{each toy graph $G^{\mathcal{T}} \in G^\mathcal{T}_{\text{TopK}}$}
                    \State Introduce noise $G^{\mathcal{T}} \leftarrow$ \textsc{AddNoise}($G^{\mathcal{T}}$) \Comment{Inner-Toy-Graph Noise}
                \EndFor
                \State Add $bottomK$ toy graphs to $G^\mathcal{T}_{\text{TopK}}$ \Comment{Toy-Graph Noise}
            \EndIf
            \State $h_c \leftarrow$ \Call{Propagation}{$G^{\mathcal{Q}}$, $G^\mathcal{T}_{\text{TopK}}$}  \Comment{Propogate hidden embedding}
            \State $o_c \leftarrow$ \Call{Propagation}{$G^{\mathcal{Q}}$, $G^\mathcal{T}_{\text{TopK}}$}  \Comment{Propogate task-specific output vector}
            
            \State Aggregated outputs $\hat{o}_c \leftarrow$ \Call{KnowledgeFusion}{$h_c,o_c$}
            \State Compute loss $\mathcal{L}_{\text{prompt}}$ via $\tilde{o}_c$ and $\hat{o}_c$  \Comment{Based on task-specific loss function}
            \State Update fine-tuned parameters $\Theta$ by minimizing $\mathcal{L}_{\text{prompt}}$
        \EndFor
    \Else
        \For{each query graph $G^{\mathcal{Q}} \in \mathcal{G}_{\text{test}}$} \Comment{Toy Graph Retrieval and Propagation}
            \State $G^\mathcal{T}_{\text{TopK}} \leftarrow$ \Call{RetrieveToyGraphs}{$G^{\mathcal{Q}}$, $\mathcal{G}^\mathcal{T}$, $TopK$}
            \State $h_c \leftarrow$ \Call{Propagation}{$G^{\mathcal{Q}}$, $G^\mathcal{T}_{\text{TopK}}$}  \Comment{Propogate hidden embedding}
            \State $o_c \leftarrow$ \Call{Propagation}{$G^{\mathcal{Q}}$, $G^\mathcal{T}_{\text{TopK}}$}  \Comment{Propogate task-specific output vector}
            \State Aggregated outputs $\hat{o}_c \leftarrow$ \Call{KnowledgeFusion}{$h_c,o_c$}
        \EndFor
    \EndIf

    \State \Return Aggregated outputs $\tilde{o}_c$
\end{algorithmic}
\end{algorithm}

\section{Further Experiment Details}
\label{sec: Further Experiment Details}

\subsection{Datasets Statics}
\label{appendix: datasets}
\begin{table}[]
    \centering
    \caption{Statistics of the experimental datasets and summary of datasets.}
    \small
    \resizebox{\linewidth}{!}{
    \begin{tabular}{l|cccccccc}
    \toprule
    Statistics & TAOBAO & KOUBEI & AMAZON & PROTEINS & COX2 & ENZYMES & BZR \\
    \midrule
    \# Nodes per Graph & 204,168 & 221,366 & 238,735 & 39.06 & 41.22 & 32.63 & 35.75 \\
    \# Edges per Graph & 8,795,404 & 3,986,609 & 876,237 & 72.82 & 43.45 & 62.14 & 38.36 \\
    \# Density & 8.6e-4 & 3.3e-4 & 6.2e-5 & 4.8e-2 & 2.6e-2 & 5.9e-2 & 3.0e-2 \\
    \# Graphs & 1 & 1 & 1 & 1,113 & 467 & 600 & 405 \\
    \# Graph Classes & / & / & /  & 2 & 2 & 6 & 2 \\
    \# Node Features & 1 & 1 & 1 & 1 & 3 & 18 & 3 \\
    \# Node Classes & / & / & / & 3 & / & 3 & / \\ \midrule
    Snapshot Granularity & daily & weekly & weekly & / & / & / & / \\
    Task & Edge & Edge & Edge &  Node, Graph & Graph & Node, Graph & Graph \\
    Type & Dynamic & Dynamic & Dynamic & Static & Static & Static & Static \\
    Dataset Partition & Snapshot & Snapshot & Snapshot & Node, Graph & Graph & Node, Graph & Graph \\ 
    \bottomrule
    \end{tabular}
    }
    \label{table:dataset statistics}
\end{table}

We employ eight benchmark datasets for evaluation including four public static classification datasets for node- and graph-level tasks.

(1) PROTEINS \cite{borgwardt2005protein} is a collection of protein graphs, including the amino acid sequence, conformation, structure, and features such as active sites of the proteins. Each node represents a secondary structure, while each edge illustrates the neighboring relationship either within the amino acid sequence or in 3D space. The nodes are divided into three categories, and the graphs are classified into two distinct classes. This dataset is used for node and graph classification tasks, containing 1,113 graphs with an average of 39.06 nodes and 72.82 edges per graph, with a density of 4.8e-2.

(2) COX2 \cite{nr} is a dataset of molecular structures, including 467 cyclooxygenase-2 inhibitors. Each node represents an atom, and each edge signifies a chemical bond between atoms, such as single, double, triple, or aromatic bonds. All the molecules belong to two categories. This dataset is used for graph classification tasks, with each graph having an average of 41.22 nodes and 43.45 edges and a density of 2.6e-2.

(3) ENZYMES \cite{wang2022faith} is a dataset of 600 enzymes collected from the BRENDA enzyme database. These enzymes are labeled into 6 categories according to their top-level EC enzyme classification. This dataset is used for node and graph classification tasks, with each graph having an average of 32.63 nodes and 62.14 edges and a density of 5.9e-2.

(4) BZR \cite{nr} is a collection of 405 ligands for the benzodiazepine receptor. Each ligand is represented by a graph, and all ligands are categorized into two groups. This dataset is used for graph classification tasks, with each graph having an average of 35.75 nodes and 38.36 edges and a density of 3.0e-2.

Additionally, we leverage three publicly available datasets encompassing a wide array of real-world scenarios in dynamic recommendation (link prediction):

(5) The TAOBAO dataset captures implicit feedback data from Taobao.com, a prominent Chinese e-commerce platform, collected over a span of 10 days. This dataset is used for edge classification tasks, containing 204,168 nodes and 8,795,404 edges, with a density of 8.6e-4.

(6) The KOUBEI dataset records 9 weeks of user interactions with nearby stores on Koubei, a platform integrated within Alipay. This dataset is used for edge classification tasks, containing 221,366 nodes and 3,986,609 edges, with a density of 3.3e-4.

(7) The AMAZON dataset comprises a collection of product reviews sourced from Amazon, spanning a duration of 13 weeks. This dataset is used for edge classification tasks, containing 238,735 nodes and 876,237 edges, with a density of 6.2e-5.

These datasets' detailed statistics are summarized in Table \ref{table:dataset statistics}. The "Task" column provides information about the type of downstream task conducted on each dataset: "Node" denotes node classification tasks, "Graph" signifies graph classification tasks, and "Edge" indicates tasks related to link prediction. The "Type" column indicates the type of graph dataset: "Dynamic" for dynamic dataset $t\ge 1$, and "Static" for static dataset $t=1$. For dynamic datasets, the "Snapshot Granularity" denotes the time granularity for each dataset. In our experimental setup, for dataset partition, dynamic graphs are partitioned according to snapshots, while static graphs are partitioned either by node or by the entire graph.

\subsection{Evaluation Matrices}
\label{appendix:evaluation matrics}
\textbf{Node and Graph classification evaluation.} For the node classification, we use the prediction accuracy to measure the model.

\textbf{Link prediction evaluation.} For the link prediction, 
we evaluate the recall and ranking quality of the effects of recommendation following previous studies \cite{simgcl,he2020lightgcn}.
We use Recall@k and NDCG@k as metrics. 
Note that this task should be a binary task. We denote the $topk$ largest value as $rel_{ij}, j\in[1,k]$ for node $v_i$.

Recall@k measures the ratio of true positive links contained in the top $k$ predicted links for each node:
\begin{equation}
\text{Recall}@k = \frac{1}{n} \sum_{i=1}^n 
\sum_{j=1}^k \frac{rel_{ij}}{\sum \mathbb{I}(A[i:]>0)},
\label{eq:recall}
\end{equation}
where $rel_{ij}=1$ if the $j$-th predicted link for node $v_i$ exists, otherwise 0.  $\mathbb{I}(\cdot)$ is the indicator function, and if $A[i:]>0$ then $\mathbb{I}(A[i:]>0)=1$.

NDCG@k (Normalized Discounted Cumulative Gain) is computed by normalizing DCG@k (Discounted Cumulative Gain) which accounts for the position of correctly predicted links. DCG@k is defined as:
\begin{equation}
\text{DCG}@k = \frac{1}{n} \sum_{i=1}^n 
\sum_{j=1}^k \frac{rel_{ij}}{\log_2 (j+1)}.
\label{eq:dcg}
\end{equation}

\subsection{Baseline Details}
In this section, we present the details of baselines.
\label{sec:appendix baselines}
\begin{table*}[ht]
\centering
\caption{Baseline Code URLs of Github Repository}
\begin{tabular}{l l l}
\hline
\textbf{Baseline} & \textbf{Type} & \textbf{Code Repo URL} \\
\hline
GCN & Static & \url{https://github.com/tkipf/gcn} \\
GraphSAGE & Static & \url{https://github.com/williamleif/GraphSAGE} \\
GAT & Static & \url{https://github.com/PetarV-/GAT} \\
GIN & Static & \url{https://github.com/weihua916/powerful-gnns} \\
LightGCN & Dynamic &\url{https://github.com/kuandeng/LightGCN} \\
SGL & Dynamic &\url{https://github.com/wujcan/SGL-Torch} \\
MixGCF & Dynamic &\url{https://github.com/Wu-Xi/SimGCL-MixGCF} \\
SimGCL & Dynamic & \url{https://github.com/Wu-Xi/SimGCL-MixGCF} \\
GraphPro &Dynamic & \url{https://github.com/HKUDS/GraphPro}
\\
GraphPrompt & Dynamic, Static & \url{https://github.com/Starlien95/GraphPrompt}
\\ 
PRODIGY & Dynamic, Static & \url{https://github.com/snap-stanford/prodigy} \\
\hline
\end{tabular}
\label{tab:repo}
\end{table*}

\begin{itemize}[leftmargin=*]
    \item \textbf{GCN}~\cite{GCN}: GCN is an end-to-end learning framework for graph-structured data. It utilizes neighborhood aggregation to integrate structural information, which is particularly effective in node classification and graph classification tasks. \\ 
    \item \textbf{GraphSAGE}~\cite{hamilton2017inductive}: GraphSAGE, is a general and inductive framework that leverages node feature information (e.g., text attributes) to efficiently generate node embeddings for previously unseen data.
    \\
    \item \textbf{GAT}~\cite{GAT}: GAT is a spatial domain method, which aggregates information through the attention-learned edge weights. \\
    \item \textbf{GIN}~\cite{gin}: GIN utilizes a multi-layer perceptron to sum the results of GNN and learns a parameter to control residual connection. \\
    \item \textbf{LightGCN}~\cite{he2020lightgcn}: LightGCN learns user and item embeddings by linearly propagating them on the user-item interaction graph, and uses the weighted sum of the embeddings learned at all layers as the final embedding. \\
    \item \textbf{SGL}~\cite{sgl}: SGL is to supplement the classical supervised task of recommendation with an auxiliary self-supervised task, which reinforces node representation learning via self-discrimination. \\
    \item \textbf{MixGCF}~\cite{huang2021mixgcf}: MixGCF generates the synthetic negative by aggregating embeddings from different layers of raw negatives' neighborhoods to perform collaborative filtering. \\
    \item \textbf{SimGCL}~\cite{simgcl}: SimGCL applies unsupervised contrastive learning to enhance representation learning, making it suitable for link prediction tasks. It is applied to dynamic graphs to test its adaptability and performance.
    \item \textbf{GraphPro}~\cite{yang2024graphpro}: GraphPro extends GraphPrompt by introducing spatial and temporal prompts tailored for dynamic graph learning, enhancing the ability to capture both structural and temporal relationships within graph data.
    \item \textbf{GraphPrompt}~\cite{graphprompt}: GraphPrompt integrates pre-training and downstream tasks using a unified template approach and employs task-specific prompts to enhance sub-task learning, applicable to both dynamic and static graph contexts.
    \item \textbf{PRODIGY}~\cite{mishchenko2023prodigy}: PRODIGY focuses on facilitating downstream tasks through in-context examples and learning from the \(X \to Y\) paradigm. It is implemented to enhance learning in both dynamic and static graphs by leveraging contextual learning strategies.
\end{itemize}

\subsection{Implementation Details}
\label{app.setting-parameter}

Implementations are done using the PyTorch 2.3.0 framework~\cite{paszke2019pytorch} in Python 3.11, on an Ubuntu server equipped with 1 V100 GPU and an Intel(R) Xeon(R) CPU.

In node and graph classification tasks: 
For baseline GCN \cite{GCN}, we employ a 2-layer architecture and set the hidden dimension as 256.
For GraphSAGE \cite{hamilton2017inductive}, we utilize the mean aggregator and employ a 2-layer architecture. The hidden dimension is also set to 256.
For GAT \cite{GAT}, we employ a 2-layer architecture and set the hidden dimension as 256. Besides, we apply 8 attention heads in the first GAT layer.
Similarly, for GIN \cite{gin}, we also employ a 2-layer architecture and set the hidden dimension as 256.
For GraphPrompt, we follow~\cite{graphprompt} to employ a 2-layer GCN as the backbone and set the hidden dimensions as 256.

In the link prediction task: For LightGCN, SGL, MixGCF, SimGCL and GraphPro, we employ a 3-layer GNN architecture and set the hidden dimension as 64 with Low-Rank Adaptation (LoRA)~\cite{hu2021loralowrankadaptationlarge} rank equals to 16. For GraphPro, the backbone graph encoder is SimGCL. 

Moreover, for all three tasks, the hyper-parameters of baselines are based on the recommended values provided in the paper. In PRODIGY and \methodname, $k$ is set to 2, $topK$ is set to 5, $\gamma$ is set to 0.8 for PROTEINS and 0.5 for ENZYMES in node level, $\gamma$ is set to 0.5 for PROTEINS, 0.6 for COX2, 0.8 for ENZYMES and 0.5 for BZR in graph level, $\alpha=\lambda=0.5, K=3, w_1=w_2=w_3=0.05, w_4=0.85$. 

\subsection{Resource Graph Scalability Study}
\begin{figure*}[htp]
    \centering
    \begin{subfigure}{0.44\textwidth}
        \centering
        \includegraphics[width=\textwidth]{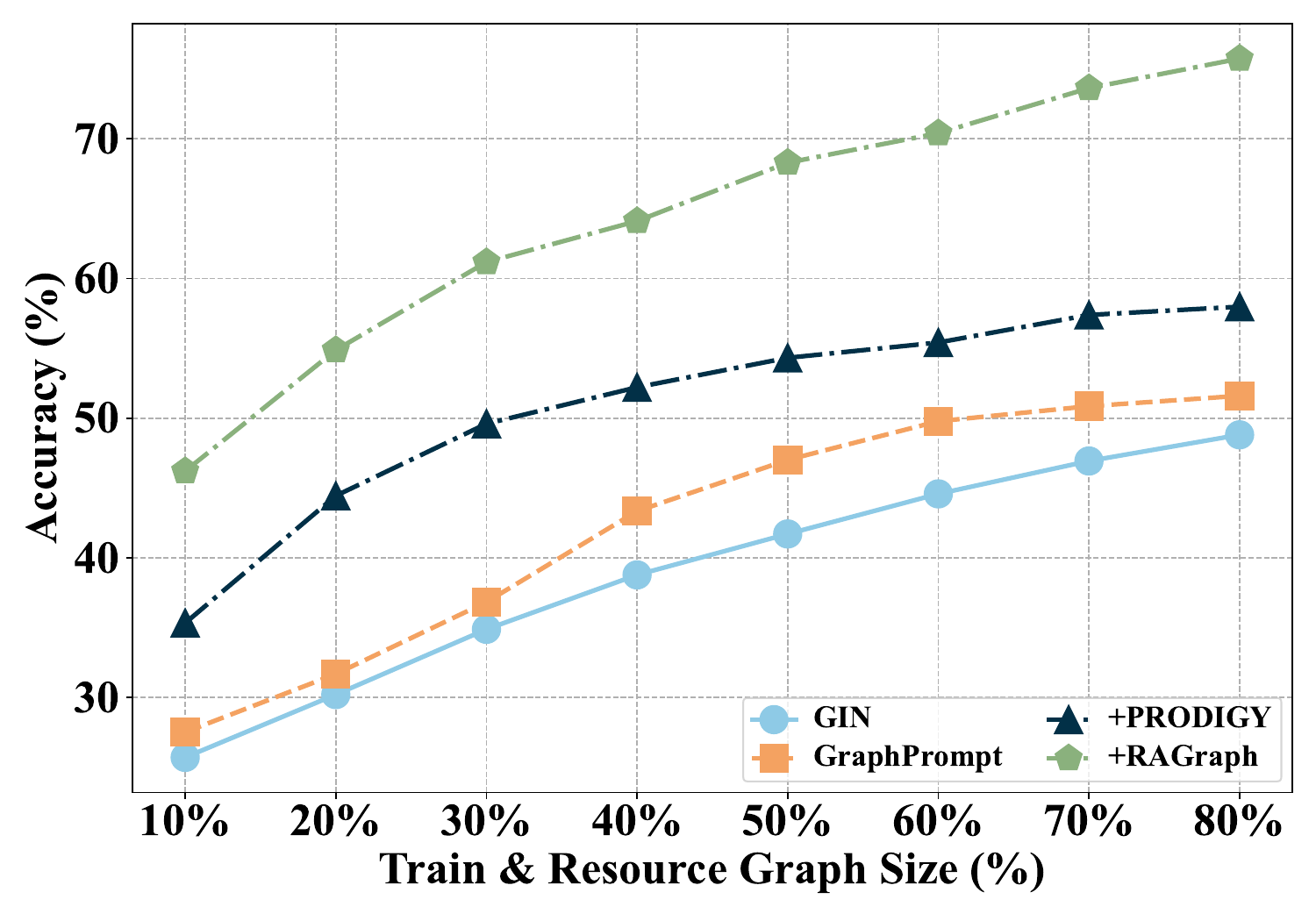}
        \caption{Node Classification on ENZYMES dataset}
        \label{fig:enzymes-node}
    \end{subfigure}
    \begin{subfigure}{0.44\textwidth}
        \centering
        \includegraphics[width=\textwidth]{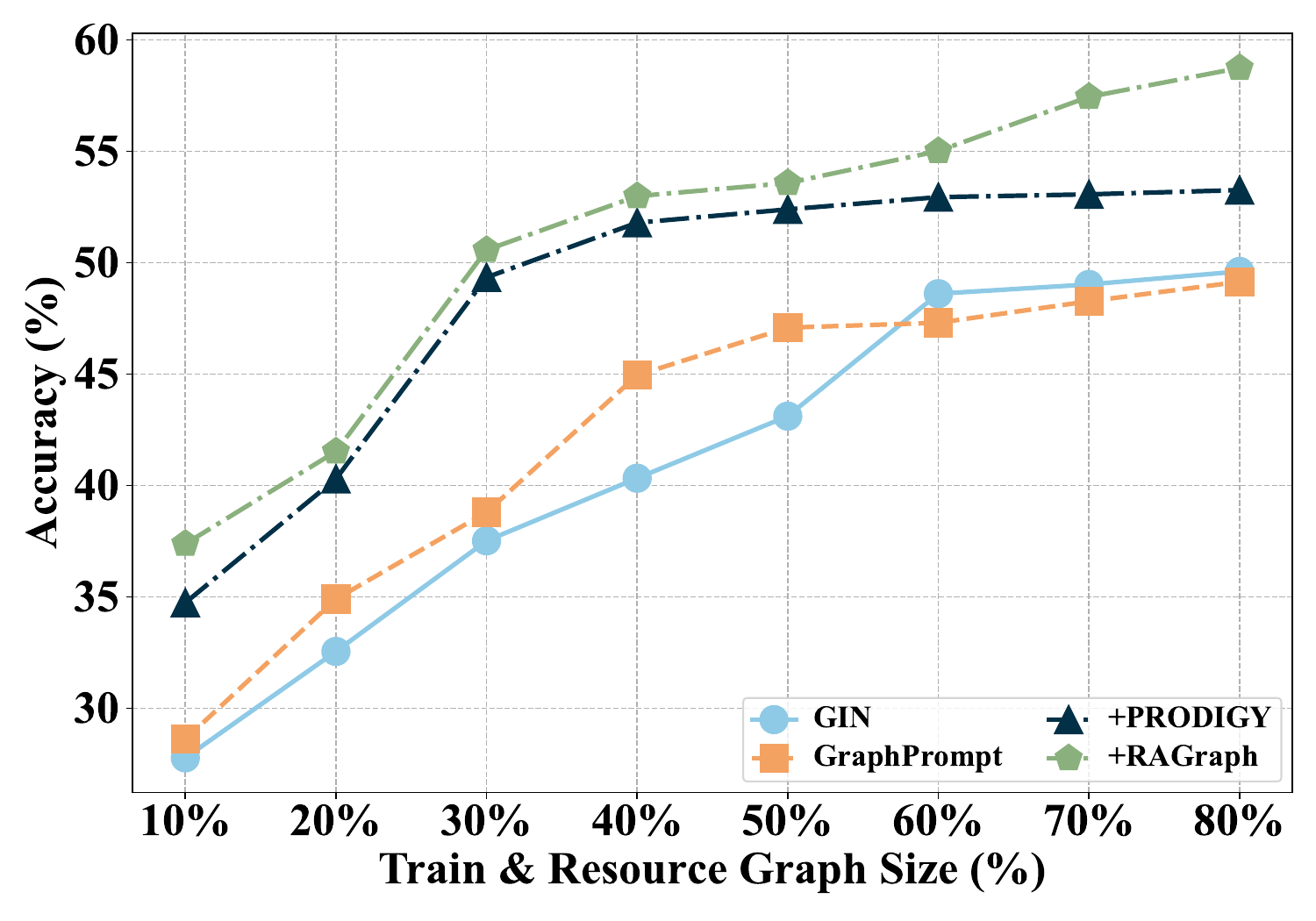}
        \caption{Node Classification on PROTEINS dataset}
        \label{fig:proteins-node}
    \end{subfigure}
    \caption{Performance comparisons of \methodname and several
baselines with different proportions of training and resource data.}
    \label{fig:scale exp}
\end{figure*}
We assess the impact of varying amounts of training and resource data on model performance. As illustrated in Figure~\ref{fig:scale exp}, we vary the proportion of train and resource graph size from 10\% to 80\%, with increments of 10\%, and conduct experiments on node classification tasks using the ENZYMES and PROTEINS datasets, respectively. For comparative analysis, we select GIN, GraphPrompt, and PRODIGY as baseline models. To ensure fairness in our experiments, we maintain a consistent ratio of train to resource data at 3:5 during fine-tuning, utilizing the sum of these as a retrieval database.

As shown in Figure~\ref{fig:scale exp}, there is a clear trend where the accuracy of the model improves as the proportion of the dataset increases. However, the rate of accuracy improvement starts to plateau once a certain dataset proportion is reached (\emph{i.e.}, 30\% in PROTEINS for PRODIGY and \methodname, 40\% in PROTEINS for GraphPrompt). Among the evaluated models, GIN and GraphPrompt show the slowest convergence rates, whereas PRODIGY converges at a moderate pace, and \methodname converges the fastest. This rapid convergence in PRODIGY and \methodname is attributed to its ability to engage in effective knowledge retrieval, significantly enhancing the model’s comprehension abilities. Remarkably, both PRODIGY and \methodname can achieve commendable results in downstream tasks even with a small proportion of the dataset. Compared to PRODIGY, \methodname exhibits superior performance because while PRODIGY primarily learns a mapping from \(X\) to \(Y\), \methodname not only learns this mapping but also integrates additional knowledge into GNNs more effectively. This integration becomes increasingly beneficial as the dataset proportion grows, allowing \methodname to outperform other models, particularly at higher data volumes where it can better leverage its knowledge integration capabilities.

\subsection{Qualitative Analyses of Toy Graphs Retrieving}
\label{app:works}

In this section, we conduct qualitative analyses of the toy graphs retrieving experiment. For the sake of understanding, we conduct experiments under normal settings where the dimensionality of the task-specific output vector is equal to the number of classes.

On the ENZYMES dataset, for a 3-class node classification task, regarding node "13984", which belongs to class 3, if we only use the GraphPrompt Backbone, the resulting one-hot encoding is $[0.28, 0.34, 0.38]$.

However, since the node is of class 3, we expect the one-hot encoding to be as close as possible to $[0,0,1]$. In \methodname retrieval, taking the top 3 retrieved graphs as examples, the connection weights for these 3 toy graphs to query graphs are 0.5, 0.7, and 0.1, respectively, and their corresponding label one-hot encodings are $[0,0,1]$, $[0,0,1]$, and $[0,1,0]$. Therefore, the result obtained by propagating the task-specific output vector through toy graphs is: $[0, 0.1, 1.2]$, and after normalization, the result is $[0, 0.08, 0.92]$.

Meanwhile, the vector obtained by propagating toy graphs hidden embedding and via decoder is $[0.37, 0.32, 0.66]$.
The retrieval of toy graphs notably enhances performance at both the task-specific output vector and hidden embedding levels. The final vector is obtained through a weighted sum with $\gamma=0.5$ in Eq(6) is $[0.185, 0.20, 0.79]$, after normalization the result is $[0.157, 0.170, 0.673]$, which greatly enhances the model's discriminative ability compared to GraphPrompt $[0.28, 0.34, 0.38]$.

\begin{figure*}[!htb]
    \centering
    \includegraphics[width = 0.8\linewidth]{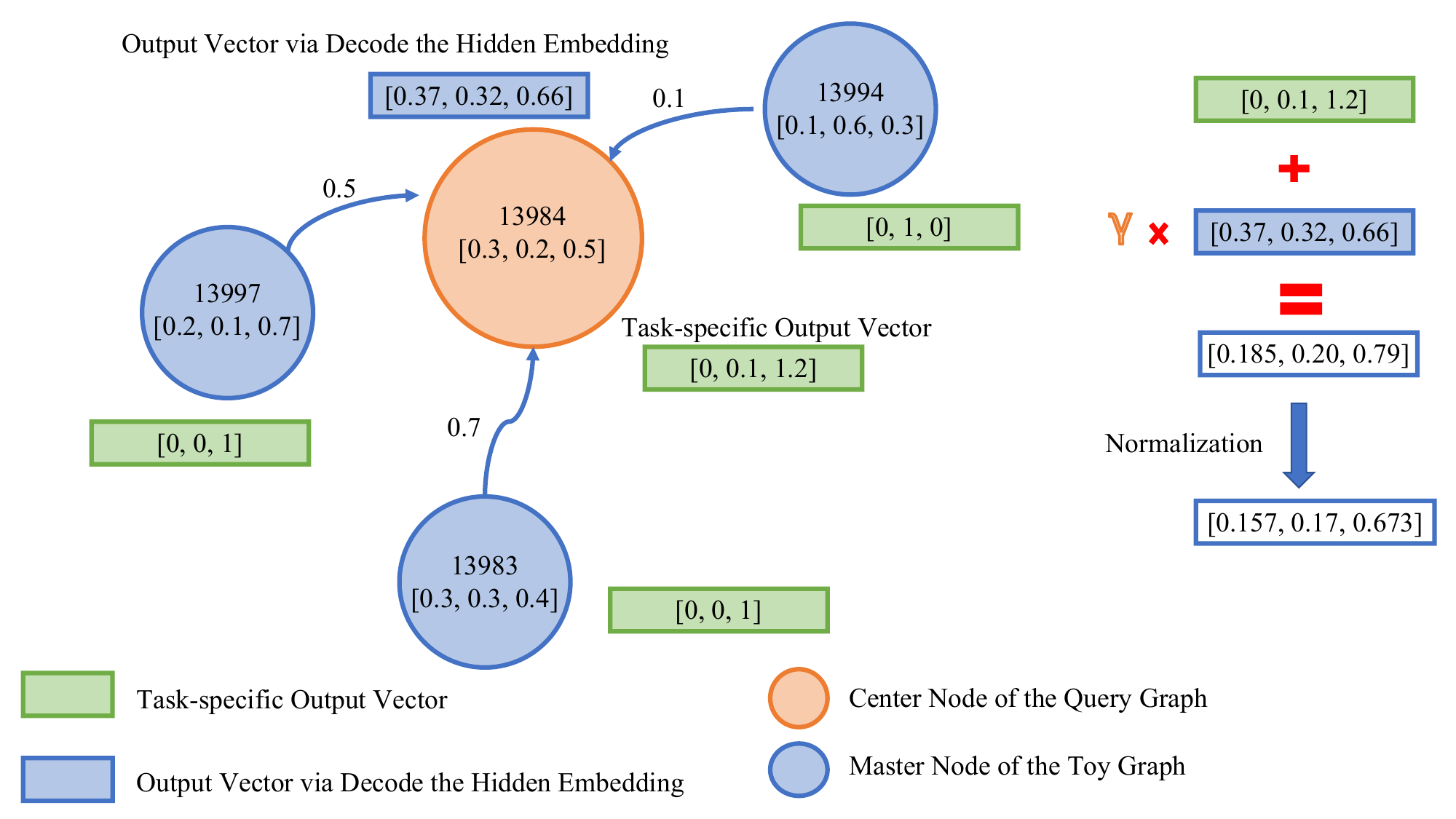}
    \caption{
        Qualitative analyses of toy graphs retrieving -- how “generation” works.
    }
\label{fig:appendix_figure1}
\end{figure*}

\section{Difference between ICL (PRODIGY) and RAG (\methodname)}
\label{sec: app diference}

\begin{figure}[tbh!]
    \centering
    \includegraphics[width = 1.0\linewidth]{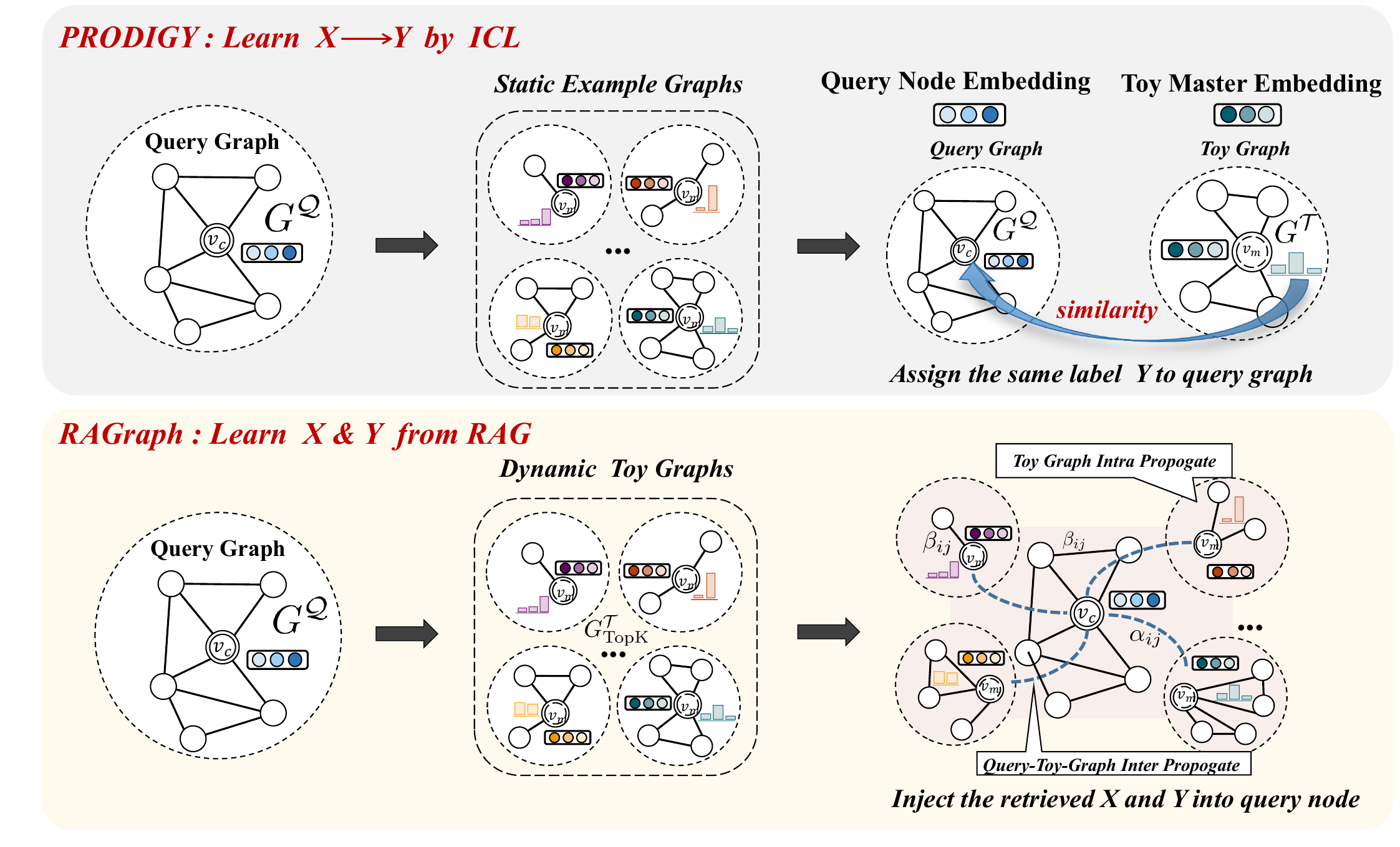}
    \caption{
        Difference Illustration between {PRODIGY} and {\methodname}. 
    }
\label{fig:appendix_figure}
\end{figure}

In this section, we explore the distinctions between PRODIGY and \methodname from several critical perspectives, as illustrated in Figure~\ref{fig:appendix_figure}:
\begin{itemize}[leftmargin=*]
    \item \textbf{PRODIGY:} This approach utilizes fixed examples as rules, which may not be optimal for dynamic and evolving scenarios. PRODIGY primarily focuses on learning direct mappings from \(X\) to \(Y\) through in-context learning. 
    However, it encounters challenges in integrating external information that is more pertinent to the query node. 
    This is particularly problematic when the distribution of each node belonging to the same label class varies, and simply learning the mapping based on the prototype node will somehow be misleading.
    \item \textbf{\methodname:} In contrast, \methodname is designed to handle non-static, streaming knowledge, making it well-suited to dynamic graph structures and evolving tasks. It actively retrieves relevant knowledge on-demand, effectively incorporating information about both \(X\) and \(Y\) from external sources into GNNs. Moreover, \methodname can operate without the need for model fine-tuning, providing substantial flexibility. This adaptability enables \methodname to excel in tasks that require continuous adaptation to changing conditions and the integration of external, relevant information.
\end{itemize}

In summary, we argue that a qualified Retrieval-Augmented Generation (RAG) system for Graph Learning should fulfill several essential criteria to effectively support complex reasoning tasks: 1) It should retrieve ample feature and task-related label information, analogous to how attributes are gathered in the NLP domain to stimulate the reasoning capabilities of LLMs; 2) The system should adapt to new tasks or unseen datasets without requiring fine-tuning of model parameters; 3) Knowledge within the system must be dynamically updated and stored, ensuring current and relevant data utilization.

\section{Broader Impacts}
\label{sec: bordader}
Our work builds on the widespread application of Retrieval-Augmented Generation (RAG) in large language models (LLMs) and aims to extend its success to graph data, thereby constructing graph foundation models. This approach allows models to transfer rapidly without requiring learnable parameters, avoiding potential performance degradation from fine-tuning pre-trained models. As a result, RAG is particularly effective in domains with scarce and long-tail data, such as network anomaly detection, rare disease diagnosis/treatment, supply chain disruption, and new user recommendations. Additionally, our model establishes an excellent paradigm by incorporating retrieved features and label information into the learning process, significantly enhancing the model's understanding capabilities. Our work provides valuable insights and serves as a reference for future Large Graph Models.

\section{Data Ethics Statement}
To evaluate the efficacy of this
work, we conducted experiments that only use publicly available
datasets, namely, PROTEINS, COX2, ENZYMES, BZR\footnote{\url{https://chrsmrrs.github.io/datasets/}}, TAOBAO, KOUBEI and AMAZON in accordance to their usage terms
and conditions if any.
We further declare that no personally identifiable information was
used, and no human or animal subject was involved in this research.

\end{document}